\documentclass[10pt,twocolumn,letterpaper]{article}

\usepackage[pagenumbers]{cvpr} 

\usepackage{graphicx}
\usepackage{amsmath}
\usepackage{amssymb}
\usepackage{booktabs}
\usepackage{tabularx,booktabs}
\usepackage{enumitem}
\usepackage[font=small,skip=1pt]{caption}
\usepackage{multirow}

\DeclareMathOperator*{\argmax}{arg\,max}
\DeclareMathOperator*{\argmin}{arg\,min}

\makeatletter
\def\thickhline{%
  \noalign{\ifnum0=`}\fi\hrule \@height \thickarrayrulewidth \futurelet
   \reserved@a\@xthickhline}
\def\@xthickhline{\ifx\reserved@a\thickhline
               \vskip\doublerulesep
               \vskip-\thickarrayrulewidth
             \fi
      \ifnum0=`{\fi}}
\makeatother

\newlength{\thickarrayrulewidth}
\usepackage[pagebackref,breaklinks,colorlinks]{hyperref}

\usepackage[capitalize]{cleveref}
\crefname{section}{Sec.}{Secs.}
\Crefname{section}{Section}{Sections}
\Crefname{table}{Table}{Tables}
\crefname{table}{Tab.}{Tabs.}

\newcommand\blfootnote[1]{%
  \begingroup
  \renewcommand\thefootnote{}\footnote{#1}%
  \addtocounter{footnote}{-1}%
  \endgroup
}

\def\cvprPaperID{xxxx} 
\def\confName{CVPR}
\def\confYear{2022}

\begin{document}

\title{Omni-DETR: Omni-Supervised Object Detection with Transformers}

\author{Pei Wang$^{2,\star}$ \ \ Zhaowei Cai$^{1,\dagger}$ \ \ Hao Yang$^1$ \ \ Gurumurthy Swaminathan$^1$ \\ Nuno Vasconcelos$^2$ \ \ Bernt Schiele$^1$ \ \ Stefano Soatto$^1$\\
AWS AI Labs$^1$\ \ UC San Diego$^2$\\
{\tt\small \{zhaoweic,haoyng,gurumurs,bschiel,soattos\}@amazon.com \ \ \{pew062,nuno\}@ucsd.edu}
}


\maketitle

\begin{abstract}
We consider the problem of omni-supervised object detection, which can use unlabeled, fully labeled and weakly labeled annotations, such as image tags, counts, points, etc., for object detection. This is enabled by a unified architecture, Omni-DETR, based on the recent progress on student-teacher framework and end-to-end transformer based object detection. Under this unified architecture, different types of weak labels can be leveraged to generate accurate pseudo labels, by a bipartite matching based filtering mechanism, for the model to learn. In the experiments, Omni-DETR has achieved state-of-the-art results on multiple datasets and settings. And we have found that weak annotations can help to improve detection performance and a mixture of them can achieve a better trade-off between annotation cost and accuracy than the standard complete annotation. These findings could encourage larger object detection datasets with mixture annotations. 
The code is available at \url{https://github.com/amazon-research/omni-detr}. 
\blfootnote{$^\star$ Work done during internship at Amazon. $^\dagger$Corresponding author.}
\end{abstract}

\section{Introduction}
\label{sec:intro}

Most of the successes of recent object detection are attributed to the large-scale well-established object detection datasets~\cite{everingham2010pascal,lin2014microsoft,shao2019objects365,dollar2009pedestrian,kuznetsova2018open}, which have accurate and complete detection annotation (category and bounding box or segmentation mask) for every object of interest in an image. In general, complete and accurate detection annotation is very expensive. For example, complete annotation of a single image of MS-COCO \cite{lin2014microsoft} takes about $346$ seconds, $76.5$ seconds on each category elimination and $269.5$ seconds on accurate bounding box localization, on average~\cite{ren2020ufo}\footnote{Not accurate numbers from \cite{lin2014microsoft} but rough estimation from \cite{ren2020ufo}.}.
Given this expensive cost, it is very difficult to scale up the data size. For example, OpenImages consisting of $9$ million images \cite{kuznetsova2018open} used a combination of machine annotation and human verification to reduce the annotation cost. 
The question is, \textit{do we need accurate and complete annotation which is expensive to achieve strong detection performances?}

\begin{figure}
  \centering
  \setlength{\tabcolsep}{2pt}
  \begin{tabular}{cc}
    \multicolumn{2}{c}{\includegraphics[width=0.98\linewidth]{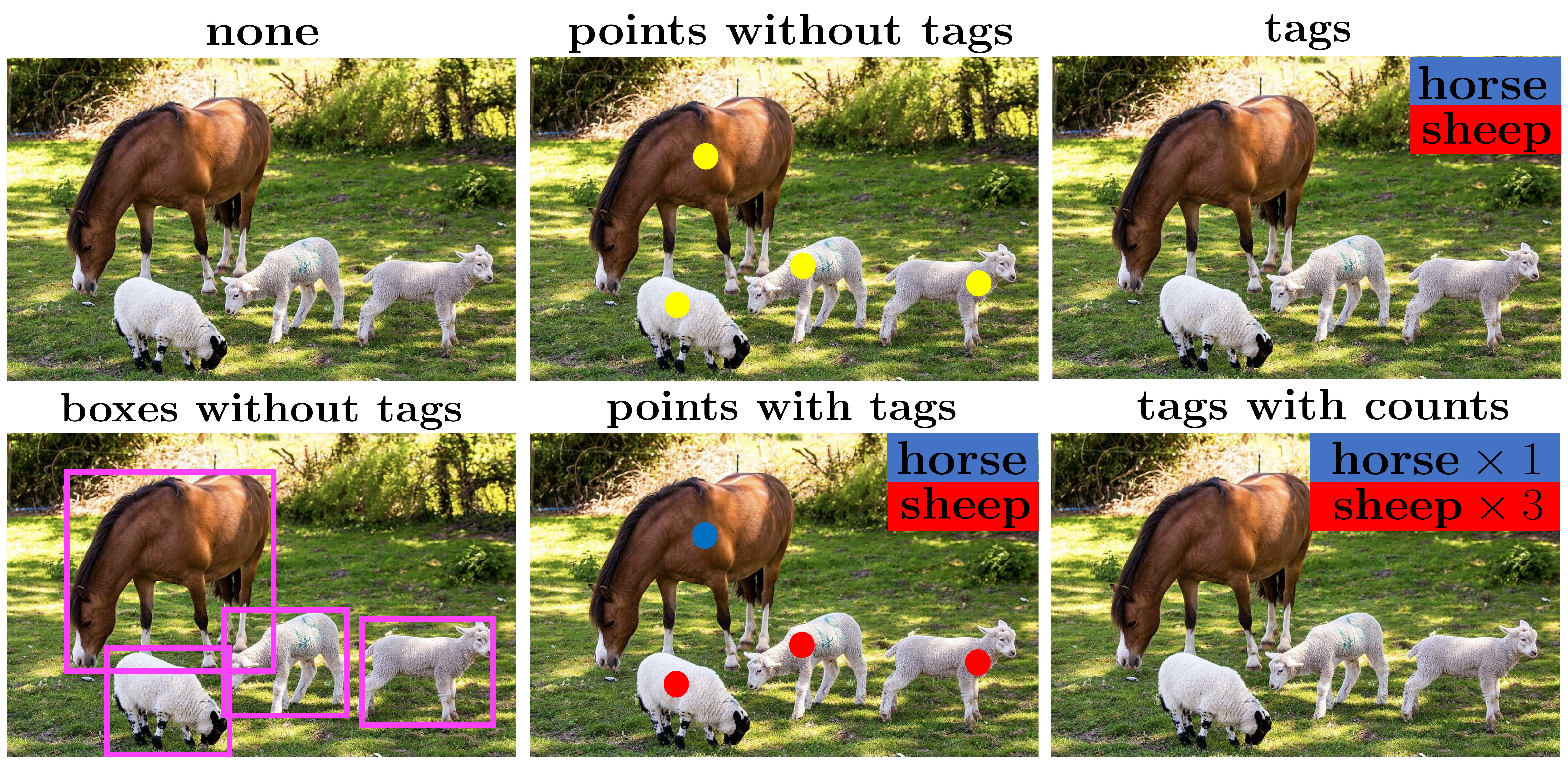}}\\
    \includegraphics[width=0.49\linewidth]{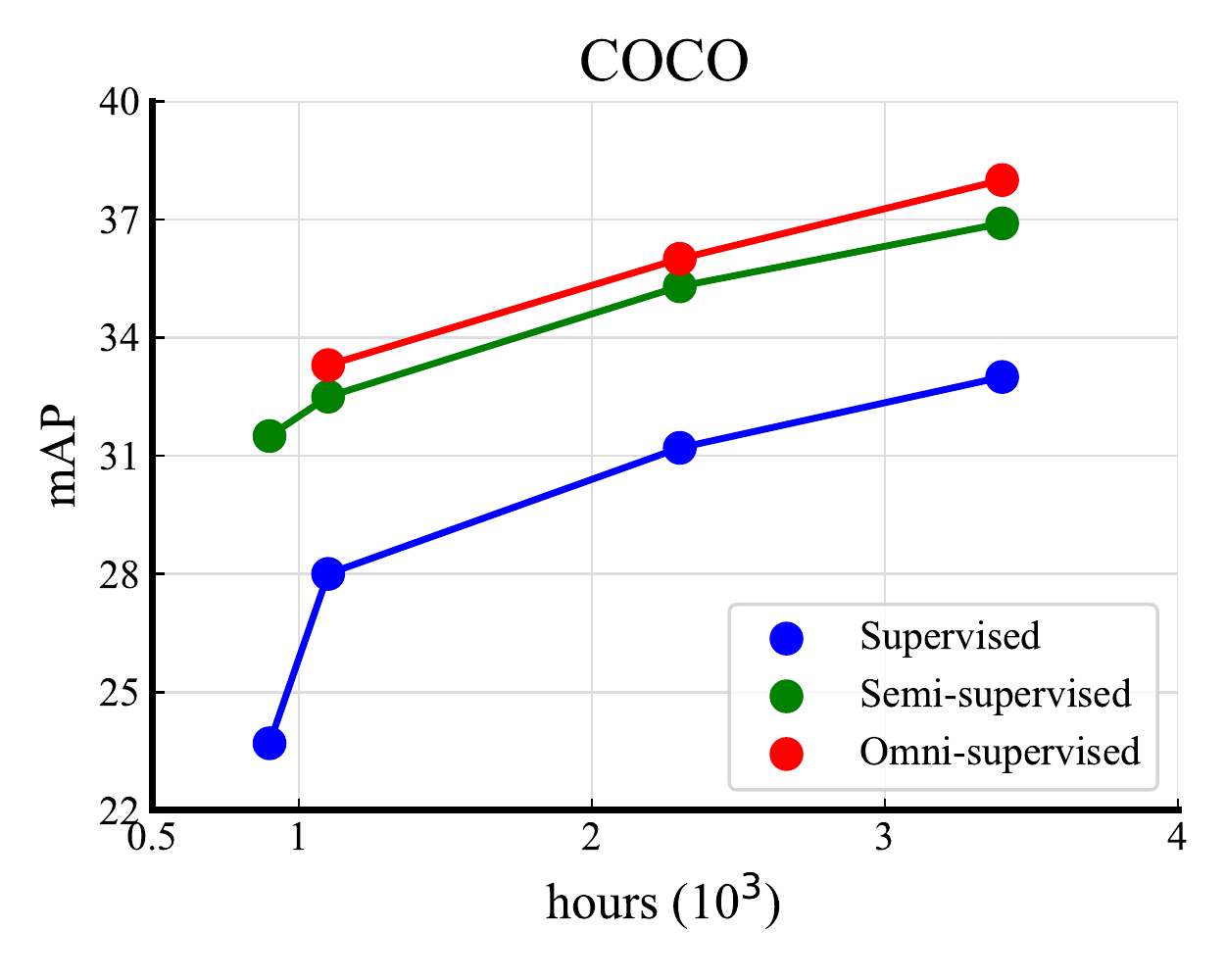} & \includegraphics[width=0.49\linewidth]{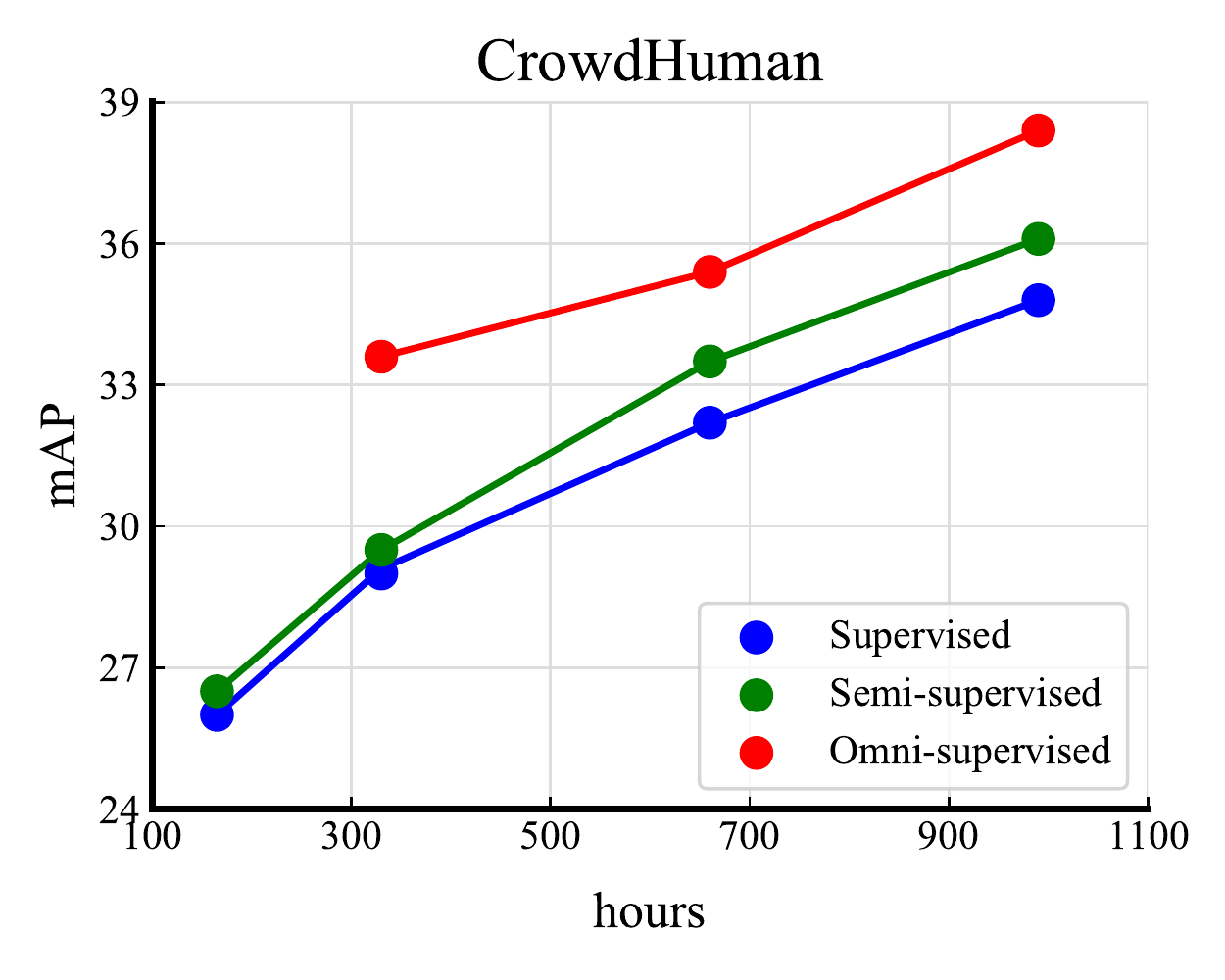}
  \end{tabular}
  \caption{The top is the visualization of different forms of weak annotations, and the bottom is the trade-off comparison (accuracy v.s. annotation cost) of supervised/semi-supervised/omni-supervised detection (see Section \ref{subsec:budge-aware detection} for more details).}
  \label{fig:teaser}
\end{figure}  

There are many weaker forms of object annotation as shown in Figure \ref{fig:teaser} (top), e.g., points, tags, counts, etc., but they are not well explored in the literature and the majority of the object detection frameworks are designed to be used with complete detection annotations. One of the main reasons for this is that using weaker forms of annotation has not shown promising results yet. For example, the performance of weakly supervised object detection (WSOD) \cite{tang2018pcl,jie2017deep,papadopoulos2017training} lags in performance compared to standard supervised detection using complete annotations. In addition,  UFO$^2$~\cite{ren2020ufo}, as the first work in omni-supervised object detection (OSOD), has shown that using additional weak annotations only has marginal gains. 
In this paper, however, we will show that, \textit{weak annotation can help to improve detection performance and achieve better cost-accuracy trade-off.}

Towards this, we propose a unified architecture for OSOD, Omni-DETR, which can work with different types of weak annotations, including image tags, object counts, points, loose bounding boxes without tags, etc., or a mixture of them. It is built on recent progresses on student-teacher based semi-supervised object detection (SSOD) \cite{liu2021unbiased,sohn2020simple,Tang_2021_CVPR} to better leverage the data even if it is unlabeled, and the end-to-end detection architecture of \cite{carion2020end,zhu2020deformable} with no heuristic detection procedures, like proposal detection, non-maximum suppression, thresholding, etc. The weak ground truth labels are used to filter the teacher predictions to generate pseudo labels for the student to learn. We formulate the pseudo-label filtering as a bipartite matching problem between the sets of predictions and available weak ground truths, and propose a unified pseudo-label filtering strategy to accommodate any form of weak annotations.

\begin{figure}[t]
\setlength{\abovecaptionskip}{-2.0pt}
\setlength{\tabcolsep}{2pt}
\begin{center}
  \includegraphics[width=1.0\linewidth]{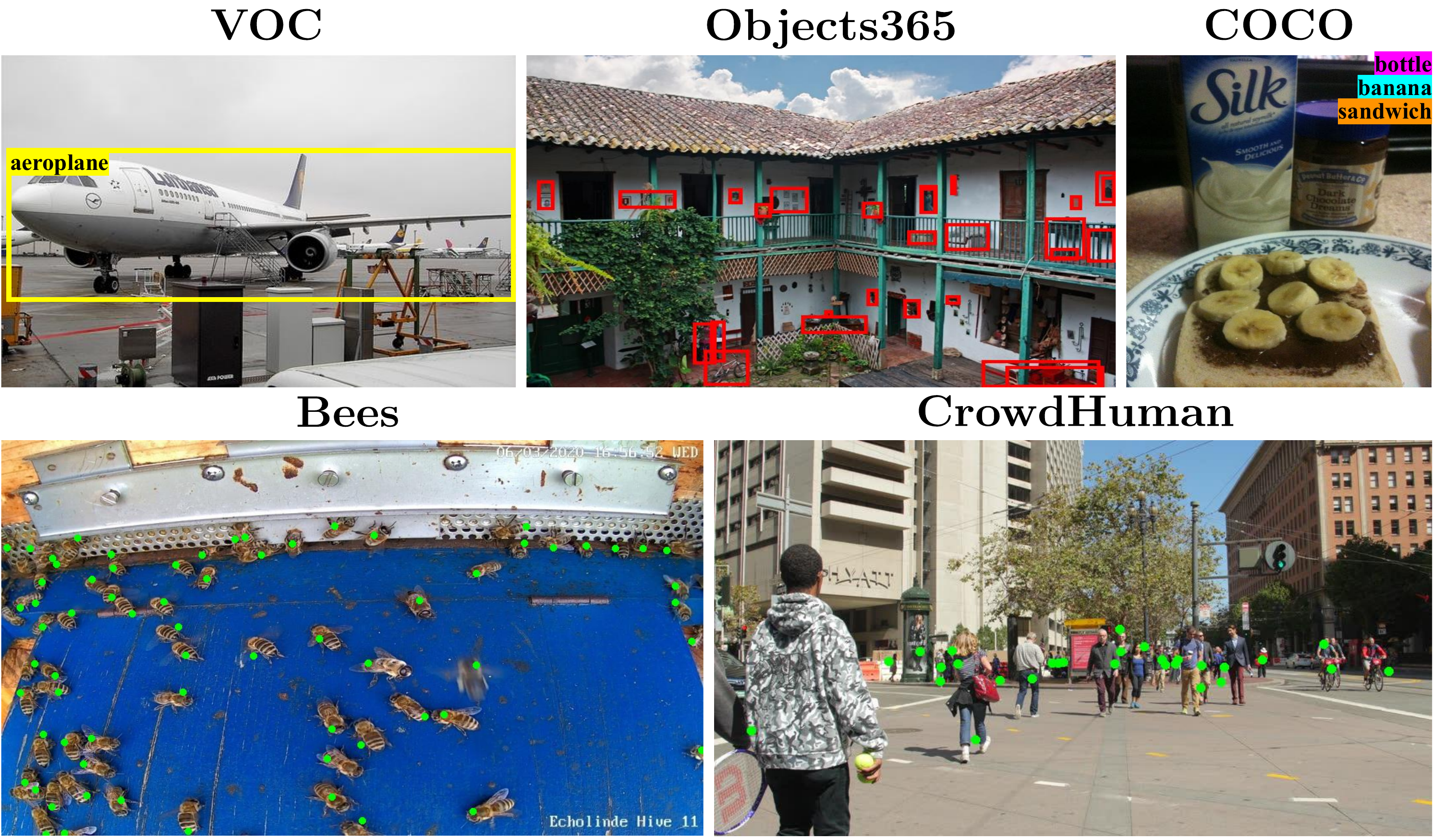}
\end{center}
  \caption{The potentially most suited annotation formats vary from dataset to dataset.}
\label{fig:dataset_annotation}
\end{figure}

Omni-DETR provides a unified framework to explore different weak forms of object annotations. 
With this framework, we have found that 1) weak annotations can bring additional gains even on a strong baseline; and 2) a mixture of weak and complete annotations can achieve better accuracy-cost trade-offs than just using complete annotations. As seen in Figure \ref{fig:teaser} (bottom), our Omni-DETR achieves better results than a standard supervised and a stronger semi-supervised detection baseline.
In addition, some annotation forms are more suited than the others depending on the dataset characteristics. For example, as shown in Figure \ref{fig:dataset_annotation}, annotating accurate bounding boxes is difficult for Bees \cite{bees} and CrowdHuman \cite{shao2018crowdhuman} datasets, since objects are small and very crowded. However, it is easier to annotate with points for these datasets. Similarly, annotating accurate categories is difficult for Objects365 \cite{shao2019objects365} as there are too many categories (365), but annotating just the bounding boxes is relatively easy and cheap.
Omni-DETR can accommodate all these different cases and help to reduce the cost of annotating such datasets, encouraging a larger scale of object detection datasets with mixed annotations.

Our contributions are summarized as: 1) a unified framework, Omni-DETR, that can accommodate various forms of object annotations or a mixture of them. 2) a novel and unified pseudo label filtering strategy, based on bipartite matching; 3) experimental findings that show weak annotations can provide additional gains and achieve better accuracy-cost trade-off than the standard full detection annotations; 4) the empirical exploration of optimal annotation mixtures for a fixed annotation budget, showing that the optimal mixture depends on the dataset.

\section{Related Work}

\textbf{Supervised Object Detection} 
is a fundamental problem in computer vision \cite{girshick2014rich,girshick2015fast,ren2015faster,lin2017feature,cai2018cascade,redmon2016you,liu2016ssd,tian2019fcos}. Most detection frameworks can be categorized into two groups: two-stage \cite{girshick2014rich,girshick2015fast,ren2015faster,lin2017feature,cai2018cascade} v.s. single-stage detectors \cite{redmon2016you,liu2016ssd,tian2019fcos}. These detectors usually have some heuristic steps, e.g., proposal detection, thresholding, non-maximum suppression, etc. More recently, \cite{carion2020end} proposed the DETR framework based on transformer \cite{vaswani2017attention} for end-to-end detection. It formulates detection as a set-to-set prediction problem, eliminating some of the previous heuristics and enabling a simpler detection pipeline. The subsequent Deformable DETR \cite{zhu2020deformable} improved on the slow training convergence of DETR and achieved better detection performances. Our Omni-DETR is also based on this end-to-end framework, which is now extended to support various forms of annotations.

\textbf{Semi-Supervised Object Detection}
(SSOD) tries to improve detection performances by using additional unlabeled data \cite{rosenberg2005semi,jeong2019consistency,sohn2020simple,liu2021unbiased,Tang_2021_CVPR,DBLP:conf/cvpr/YangWW0021,DBLP:conf/cvpr/0001YWQL21}. A prevalent SSOD paradigm is to use a multi-stage self-training pipeline \cite{rosenberg2005semi,xie2020self,sohn2020simple}: 1) train model on labeled data; 2) generate pseudo labels on unlabeled data; 3) retrain model on both labeled and pseudo-labeled data; 4) repeat this process if needed. Some recent works \cite{jeong2019consistency,liu2021unbiased,Tang_2021_CVPR} have shown great progress by resorting to an online pipeline. \cite{jeong2019consistency} leverages the consistency regularization between two different augmented views of a single image. \cite{liu2021unbiased,Tang_2021_CVPR} rely on a mean-teacher framework \cite{tarvainen2017mean}, where the teacher generates online pseudo labels for the student to learn. Omni-DETR is also based on this mean-teacher framework, but uses different weak annotations to generate accurate pseudo labels.

\textbf{Weakly-Supervised Object Detection}
(WSOD) aims to reduce detection annotation efforts by leveraging cheaper weakly labeled data. Most works only use a single type of weak annotations, e.g. image-level tags \cite{tang2018pcl,jie2017deep,gokberk2014multi,song2014weakly,bilen2016weakly} or instance-level points \cite{papadopoulos2017training,chandra2020active,gygli2019efficient},
and usually formulate WSOD as a multiple instance learning (MIL) problem. This, however, has very limited success so far. Some recent works study weakly semi-supervised object detection (WSSOD) \cite{gao2019note,fang2021wssod,chen2021points}, using a small amount of fully labeled data and a large amount of additional weakly labeled data. This has shown more promising results than WSOD. In general, different types of weak annotations require specific detection algorithms, e.g., \cite{gao2019note,fang2021wssod} for WSSOD with tags and \cite{chen2021points} for WSSOD with points. The proposed Omni-DETR is closely related to WSSOD but can accommodate various weak annotations instead of a single one.

\textbf{Omni-Supervised Object Detection}
(OSOD) combines different forms of annotations to improve detection. It was first proposed in UFO$^2$ \cite{ren2020ufo}, which is based on the Faster R-CNN~\cite{ren2015faster} framework and formulates OSOD as a multitask learning problem. However, UFO$^2$ has only shown very marginal improvements for the addition of weak annotations, and suggested that, for a fixed annotation budget, the best choice is still full annotation. However, our experiments of Omni-DETR have the opposite observations: weak annotations are helpful and a mixture of annotations is a better solution than full annotation given a fixed annotation budget. 
Table \ref{tab:omni_vs_others} summarizes the related works of object detection using different types of annotations, and our Omni-DETR is a more universal framework on annotation formats than the previous works.

\textbf{Object Detection Data Annotation} 
is known to be an expensive and tedious task \cite{everingham2010pascal,dollar2009pedestrian,lin2014microsoft,kuznetsova2018open,shao2019objects365,gupta2019lvis}, that requires annotators to choose the right category and localize the accurate bounding box for each object. For example, 
\cite{su2012crowdsourcing} reports an average annotation time of 35
seconds for a high-quality bounding box. The total estimated time with associated categories per COCO image is 346 seconds~\cite{ren2020ufo}.
This high annotation cost prevents the detection dataset from being scaled up, in terms of the number of images, classes and objects. Several strategies have been used to reduce the cost. For example, Caltech Pedestrian \cite{dollar2009pedestrian} interpolates annotations between two video frames, OpenImages \cite{kuznetsova2018open} uses machine prediction first and then human verification next, LVIS \cite{gupta2019lvis} only annotates a few positive/negative classes for an image instead of complete category annotation, etc. Other approaches try to relax the accurate bounding box annotation, by proposing to use relatively loose bounding boxes \cite{papadopoulos2017extreme} or near-center points \cite{papadopoulos2017training}. In this work, using Omni-DETR, we empirically find that accurate and complete detection annotation is not the most economical, and a mixture of annotations can achieve a better trade-off between accuracy and cost.

\section{Omni-DETR}

We at first introduce the overall framework of Omni-DETR in this section and then the unified pseudo-label filtering for various weak annotations in the next section.

\newcommand{\rr}[1]{\rotatebox{90}{#1}}
\begin{table}
\setlength{\abovecaptionskip}{-2.0pt}
\setlength{\tabcolsep}{1pt}
\small
\begin{center}
\begin{tabular}{l|c|c|c|c|c|c|c}
 & \rr{None} & \rr{Tags} & \rr{\begin{tabular}{@{}l@{}}Tags \\ w/ counts\end{tabular}} & \rr{\begin{tabular}{@{}l@{}}Points \\ w/o tags\end{tabular}} &\rr{\begin{tabular}{@{}l@{}}Points \\ w/ tags\end{tabular}} & \rr{\begin{tabular}{@{}l@{}}Boxes \\ w/o tags\end{tabular}} & \rr{Mixture}\\
\thickhline
SSOD~\cite{liu2021unbiased,Tang_2021_CVPR} & \checkmark& & & & & \\
WSOD with tags~\cite{tang2018pcl,jie2017deep,bilen2016weakly} & &\checkmark & & & & \\
WSOD with points~\cite{papadopoulos2017training,gygli2019efficient} & & & & &\checkmark & & \\
WSSOD with tags~\cite{fang2021wssod}  & &\checkmark & & & & \\
WSSOD with points~\cite{chen2021points}  & & & & &\checkmark & \\
UFO$^2$~\cite{ren2020ufo} & \checkmark& \checkmark& & &\checkmark & &\checkmark\\
{\bf Our Omni-DETR} & \checkmark& \checkmark& \checkmark& \checkmark& \checkmark& \checkmark& \checkmark
\end{tabular}
\end{center}
\caption{Summary of related works of object detection using different weak annotations.}
\label{tab:omni_vs_others}
\end{table}

\subsection{Omni-labels}
\label{subsec:omni-labels}

Omni-DETR is a unified framework to combine fully and weakly labeled data. 
It assumes the availability of a fully labeled and a weakly labeled dataset. The fully labeled dataset $\mathcal{D}^l = \{(\mathbf{x}^l_i,\mathbf{y}^l_i)\}_{i=1}^{N^l}$, where $\mathbf{x}^l_i$ is the $i$-th image and $\mathbf{y}^l_i = \{(\mathbf{b}_{i,j},c_{i,j}) \in \mathcal{R}^4 \times \{1,2,...,C\} \}_{j=1}^{B_i}$ the corresponding label, composed by $B_i$ pairs of 1) four coordinate bounding boxes  $\mathbf{b}_{i,j}$ and 2) corresponding classes $c_{i,j}$, assigns class label $c_{i,j}$  to the object localized by bounding box $\mathbf{b}_{i,j}$. The weakly labeled data, $\mathcal{D}^o= \{(\mathbf{x}^o_i,\mathbf{y}^o_i)\}_{i=1}^{N^o}$, $\mathbf{y}^o_i$ can consist of any of the annotations introduced in the following. We omit the image index $i$ in subsquent notations, for notational simplicity, and term the fully labeled dataset as \emph{labeled} and the weakly labeled dataset as \emph{omni-labeled}.

\label{sec:omni-sup}

Omni-DETR supports any of the blow annotation forms $\mathbf{y}$, or a mixture of them, as weak annotations for image $\mathbf{x}$.

\noindent\textbf{None (None)} $\mathbf{y} = \varnothing$. No annotation for image $\mathbf{x}$.

\noindent\textbf{Tags (TagsU)} 
$\mathbf{y} = \{c_j\}_{j=1}^{M}$, which is a list of image-level classes\footnote{We will also use ``tag'' to refer ``class'' interchangeably.}. $M$ is the number of tags. In the examples of Figure \ref{fig:teaser} (top), $M=2$, $c_{1}$ is ``horses'' and $c_{2}$ is ``sheep''. 

\noindent\textbf{Tags with counts (TagsK)}, $\mathbf{y} = \{(c_{j},n_{j})\}_{j=1}^{M}$, where $n_{j}$ is the count number of objects of class $c_{j}$. In Figure \ref{fig:teaser} (top), $c_{1}$ is ``horses'' and $n_{1}=1$, while $c_{2}$ is ``sheep'' and $n_{2}=3$. 

\noindent\textbf{Points without tags (PointsU)}, $\mathbf{y} = \{\mathbf{p}_j \in \mathcal{R}^2\}_{j=1}^{P}$, where $\mathbf{p}_j$ is a point annotation for an object,
e.g., the geometric center of the object or a random point inside the region of support of the object in the image, and $P$ the number of points. In Figure \ref{fig:teaser} (top), four point annotations identify four objects without class information.

\noindent\textbf{Points with tags (PointsK)}, $\mathbf{y} = \{(\mathbf{p}_j,c_{j})\in \mathcal{R}^2 \times \{1,2,...,C\}\}_{j=1}^{P}$. In addition to PointsU, the label of each point is also known. In Figure \ref{fig:teaser} (top), the points and labels for three sheeps and a horse are annotated.

\noindent\textbf{Boxes without tags (BoxesU)}, $\mathbf{y} = \{\mathbf{b}_j \in \mathcal{R}^4\}_{j=1}^{B}$. The standard bounding box annotation but removing the class information. $B$ is the number of boxes.

\noindent\textbf{Extreme Clicking Box (BoxesEC)}, $\mathbf{y} = \{\mathbf{b}_j \in \mathcal{R}^4\}_{j=1}^{B}$, where $\mathbf{b}_j$ is a box derived from the annotation of extreme points of the object. This 
was introduced in  \cite{papadopoulos2017extreme} and has much  less annotation cost ($5 \times$) but only slightly worse quality than BoxesU annotation.

\subsection{Unified Framework}

\begin{figure}[t]
\setlength{\abovecaptionskip}{-2.0pt}
\setlength{\tabcolsep}{2pt}
\begin{center}
  \includegraphics[width=1.0\linewidth]{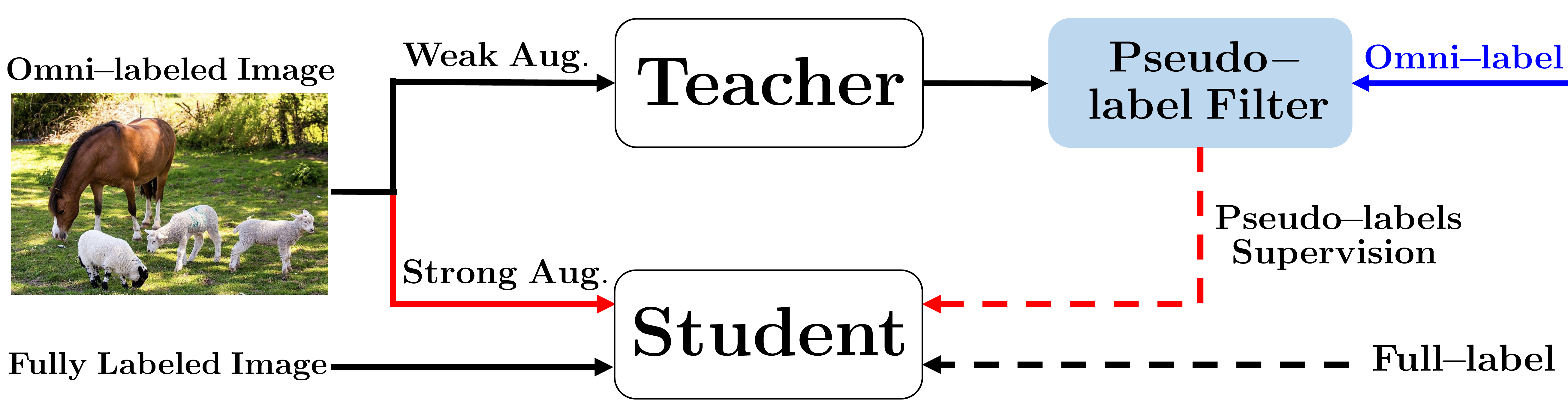}
\end{center}
  \caption{{\bf The framework of Omni-DETR}, which is based on the student-teacher framework. The {\color{blue} omni-label} is used to filter the predictions of the teacher network, by a unified pseudo-label filter, to generate pseudo-labels for the student network to learn. The omni-label can be any annotation introduced in Section \ref{subsec:omni-labels}.}
\label{fig:framework}
\end{figure}

Figure \ref{fig:framework} gives an overview of the Omni-DETR framework. Motivated by the recent successes of student-teacher frameworks for semi-supervised learning \cite{sohn2020fixmatch} and SSOD \cite{liu2021unbiased,Tang_2021_CVPR}, our Omni-DETR is also composed of a student detection network $\mathcal{F}^s(\mathbf{x}; \theta_s)$ and a teacher detection network $\mathcal{F}^t(\mathbf{x}; \theta_t)$. For the omni-labeled data $(\mathbf{x}^o, \mathbf{y}^o) \in \mathcal{D}^o$, two views of the image $\mathbf{x}^o$ are generated by a strong and a weak augmentation, $\mathbf{x}^{o,s}$ and $\mathbf{x}^{o,w}$, respectively. The weakly augmented view $\mathbf{x}^{o,w}$ is forwarded through the teacher to produce the detection prediction $\mathbf{\hat{y}}^t=\mathcal{F}^t(\mathbf{x}^{o,w}; \theta_t)$, consisting of class prediction $\mathbf{\hat{y}}^{cls}$ and bounding box prediction $\mathbf{\hat{y}}^{box}$. The predictions $\mathbf{\hat{y}}^t$ are then passed to a pseudo label filter $\mathcal{T}$ together with the available omni-labels $\mathbf{y}^o$. The filter generates
the pseudo-labels $\mathbf{\tilde{y}}^t = \mathcal{T}(\mathbf{\hat{y}}^t; \mathbf{y}^o)$,
which are used to supervise the learning of the student on the strong augmentation $\mathbf{x}^{o,s}$. The pseudo-label filtering details will be discussed in Section \ref{sec:filtering}. Here the weak/strong augmentation is only applied to teacher/student because the weak annotation can induce more accurate pseudo-labels for the teacher and the strong augmentation can make the learning of the student more challenging. For the labeled data $(\mathbf{x}^l, \mathbf{y}^l) \in \mathcal{D}^l$, strong and weak augmentations are also generated, $(\mathbf{x}^{l,s}, \mathbf{y}^{l,s})$ and $(\mathbf{x}^{l,w}, \mathbf{y}^{l,w})$, and both are feed-forwarded to the student network for learning only.

Only the student $\mathcal{F}^s$ is optimized by standard SGD with the overall loss,
\begin{equation}
    \mathcal{L}^s = \sum_i \mathcal{L}(\mathbf{x}_i^{l,s}, \mathbf{y}_i^{l,s}) + \mathcal{L}(\mathbf{x}_i^{l,w}, \mathbf{y}_i^{l,w}) + \sum_i \mathcal{L}(\mathbf{x}_i^{o,s}, \mathbf{\tilde{y}}_i^{t}),
    \label{equ:L_s}
\end{equation}
where
\vspace{-2mm}
\begin{equation}
    \mathcal{L} = \alpha \mathcal{L}^{cls} + \beta \mathcal{L}^{box}
    \label{equ:loss1}
\end{equation}
is the weighted sum of classification loss $\mathcal{L}^{cls}$ and bounding box regression loss $\mathcal{L}^{box}$, and $\alpha$ and $\beta$ are the corresponding weights. The teacher $\mathcal{F}^t$ is updated by the exponential moving average (EMA) from the student \cite{tarvainen2017mean}, 
\begin{equation}
    \theta_t \leftarrow k \theta_t + (1-k) \theta_s,
\label{equ:L_t}
\end{equation}
where $k$ is empirically set to a number close to 1, e.g., 0.9996. This EMA updated teacher can be seen as a temporal ensemble of student models along the training trajectories, which makes it more robust and able to generate more accurate pseudo-labels \cite{DBLP:conf/uai/IzmailovPGVW18,DBLP:conf/iclr/AthiwaratkunFIW19,cai2021exponential}. Note that this reduces to the Unbiased Teacher (UT) framework \cite{liu2021unbiased} proposed for SSOD when no omni-labels are available, and only unlabeled data is used. It follows that UT is a baseline for Omni-DETR and the addition of any weak annotations should improve the accuracy of this SSOD baseline. This establishes a much stronger baseline than any previous weakly supervised object detection (WSOD) and weakly semi-supervised object detection (WSSOD) work \cite{fang2021wssod,tang2018pcl,jie2017deep,papadopoulos2017training}.

\subsection{Detection Architecture}
\label{sec:detr}

Although there is no constraint on which detector to use, DETR is chosen here because it has removed many heuristic procedures in the traditional detection frameworks \cite{ren2015faster,lin2017feature,redmon2016you,liu2016ssd}. This is necessary for Omni-DETR since it needs to accommodate many different kinds of annotations. 

DETR \cite{carion2020end} is a transformer \cite{vaswani2017attention} based end-to-end object detection framework. In DETR, a standard CNN backbone 
is at first applied to a given image, and the output features are flattened and followed by an encoder transformer. In order to detect objects, the decoder transformer is applied by taking the object queries as input and cross-attending the encoded vision features, to generate the final object predictions with class and bounding box predictions $\mathbf{\hat{y}}^{cls}$ and $\mathbf{\hat{y}}^{box}$. 
Then a set-to-set alignment is enabled by using Hungarian matching \cite{kuhn1955hungarian} between the object predictions and the ground truth objects. After matching each hypothesis and ground truth, standard learning is used to optimize the classification task (with multi-class cross-entropy loss) and bounding box regression task (with generalized IoU and $L$1 loss). Due to the slow convergence of original DETR, we use Deformable DETR~\cite{zhu2020deformable} for faster convergence speeds.

\subsection{Training}

The overall model is trained with two stages: 1) burn-in training of the student network alone on the labeled data; 2) student-teacher training on both \emph{labeled} and \emph{omni-labeled} data where the teacher model is initialized by duplicating the burn-in student model.

\vspace{-2mm}
\section{Pseudo-label Filtering}
\label{sec:filtering}

As shown in Figure \ref{fig:framework}, the pseudo-label filter is a key component to leverage weak annotations in our Omni-DETR. It takes in both detection predictions and available omni-labels of an \emph{omni-labeled} image, and then generates the pseudo-labels to supervise the
learning of the student.

\subsection{Simple Pseudo-label Filtering}
\label{sec:simple_filter}

At first, we present some simple pseudo-label filtering approaches for different weak annotations.
Object detectors usually output a vector of confidence scores $\mathbf{s}_j \in [0,1]^C$ per detected bounding box ${\bf b}_j$. A popular approach to generate pseudo-labels is to simply threshold these scores. If only tag supervision (TagsU) is available, pseudo-labels can be generated by thresholding the confidence $s_j^{c_j}$ of the ground truth class $c_j$. For a ground truth class if there is no prediction greater than the confidence threshold, the top-1 prediction is retrieved as the pseudo-label for that class. When count supervision is available (TagsK), with $n_j$ counts for each ground truth class $c_j$, this can be extended to selecting the top $n_j$ predictions for class $c_j$. When point supervision is available (PointsU), a similar strategy is to choose the predicted bounding boxes that contain the ground truth points.  If additional tag supervision is available (PointsK), this can be extended to choosing candidates whose class prediction matches the point tag~\cite{ren2020ufo}. However, these empirical filtering rules are specific to each type of weak supervision and do not provide a unified pseudo-labeling solution. 

\subsection{Unified Pseudo-label Filtering}
\label{sec:unified_filtering}

Next, we introduce the proposed unified approach\footnote{Although specific design is still needed for each weak annotation, here we use ``unified'' because the filtering of different weak annotations can
be interpreted by a unified bipartite matching mechanism.}.
Formally, the filter is applied to $\mathbf{\tilde{y}} = \mathcal{T}(\mathbf{\hat{y}}; \mathbf{y}^o)$ where $\mathbf{\hat{y}} = \{\mathbf{\hat{y}}^{cls}, \mathbf{\hat{y}}^{box}\}$ is the prediction of the teacher network, with $\mathbf{\hat{y}}^{cls}$ and $\mathbf{\hat{y}}^{box}$ being the class and box predictions, respectively.
Here, we define $\mathbf{\hat{y}}^{cls} = [\mathbf{z}_1,...,\mathbf{z}_{K}]^T \in \mathcal{R}^{K \times C}$ and $\mathbf{\hat{y}}^{box} = [\mathbf{\hat{b}}_1,...,\mathbf{\hat{b}}_{K}]^T \in \mathcal{R}^{K \times 4}$, where $\mathbf{z}_k$ is a vector of logits (the network output vector before the softmax), $\mathbf{\hat{b}}_k$ the
associated bounding box prediction, and $K$ the number of object queries. $\mathbf{y}^o$ is the omni-label of section \ref{sec:omni-sup}.

\vspace{-2mm}
\subsubsection{No Annotation}
\label{sec:none_anno}

When no annotation is available (\textbf{None}), pseudo-labels are derived from confidence scores, as used in SSOD \cite{liu2021unbiased}. Specifically, $\mathbf{\hat{y}}^{cls}$
is fed into a softmax layer, to produce $[\mathbf{p}_1,...,\mathbf{p}_{K}]^T \in \mathcal{R}^{K \times C}$, where $\mathbf{p}_k$ is the probabilities over $C$ class for query $k$. 
The predicted class of the $k$-th prediction is defined as $\hat{c}_k = \argmax_c p_k^{c}$ and the confidence score as the associated probability $s_k = p_k^{\hat{c}_k}$. The threshold $\tau$ is used to filter low-confidence predictions, by collecting the prediction index set $I = \{k | s_k > \tau, k \in [1,K]\}$. With the bounding box predictions $\mathbf{\hat{y}}^{box} = [\mathbf{\hat{b}}_1,...,\mathbf{\hat{b}}_{K}]^T \in \mathcal{R}^{K \times 4}$, the pseudo-labels are then defined as $\{(\mathbf{\hat{b}}_k, \hat{c}_k) | k \in I\}$ where $\mathbf{\hat{b}}_k$ is a pseudo box and $\hat{c}_k$ its pseudo class.

\vspace{-2mm}
\subsubsection{Weak Annotations}

Motivated by DETR \cite{carion2020end}, we formulate the pseudo-label filtering problem as a bipartite matching problem, between the $K$ teacher predictions and the available $G$ ground truth omni-labels $\{\mathbf{g}_i\}^G_{i=1}$ ($G<K$).
Specifically, we search for a permutation $\hat{\sigma} \in \wp_{K}$ of $K$ elements such that
\begin{equation}
    \hat{\sigma} = \argmin_{\sigma \in \wp_{K}} \sum_i^{K} \mathcal{L}_{match} (\mathbf{g}_i,\mathbf{\hat{y}}_{\sigma(i)}),
    \label{equ:Hun_match}
\end{equation}
where $\mathcal{L}_{match} (\mathbf{g}_i,\mathbf{\hat{y}}_{\sigma(i)})$ is an annotation-specific pair-wise matching cost between ground truth omni-label $\mathbf{g}_i$ and teacher prediction $\mathbf{\hat{y}}$ of index $\sigma(i)$.
The optimal assignment is enabled with Hungarian matching~\cite{kuhn1955hungarian,carion2020end}, assigning pseudo-labels
$\{(\mathbf{b}^*_i, c^*_i)\}^G_{i=1}$. Here, $\mathbf{b}^*_i \in \mathcal{R}^4$ is the pseudo bounding box and $c^*_i$ the pseudo class, depending on the weak annotation type. 
Next, we present the specific $\mathcal{L}_{match} (\mathbf{g}_i,\mathbf{\hat{y}}_{\sigma(i)})$ and $(\mathbf{b}^*_i, c^*_i)$ for different annotations.

\vspace{-5mm}
\paragraph{TagsU} 
When the image-level ground truth tags are available, $\mathbf{y}^o = \{c_j\}^M_{j=1}$, where $M$ is the number of tags and $c_j$ is the $j$-th class, but the exact number of objects per class is not known, the matching of (\ref{equ:Hun_match}) is not directly applicable. To address this problem, the count $n_j$ of tag $c_j$, is first predicted with,
\vspace{-2mm}
\begin{equation}
    n_j = \max(1, |\{k|p_k^{c_j}>\tau,k \in [1,K]\}|),
\label{equ:tagsU}
\end{equation}
where $p_k^{c_j}$ is the probability of assigning the $k$-th prediction to class $c_j$, and $|\cdot|$ is the set cardinality. The predicted count is the number of predictions that pass the confidence threshold, if any, and set to one otherwise. This is because there is at least one object per ground truth tag.
In order to accommodate the matching of (\ref{equ:Hun_match}) regarding $G$ ground truth, we re-write the ground truth set as $\{\mathbf{g}_i\} = \{c_i\}^G_{i=1}$ with $G = \sum_j^M n_j$ and $n_j$ repetitions of each tag $c_j$. Note that, for different $i$, $c_i$ could be the same if there are multiple objects per class.
$\mathcal{L}_{match} (\mathbf{g}_i,\mathbf{\hat{y}}_{\sigma(i)})$ in (\ref{equ:Hun_match}) is defined as 
\begin{equation}
    \mathcal{L}^t_{match} (\mathbf{g}_i,\mathbf{\hat{y}}_{\sigma(i)}) = 1 - p_{\sigma(i)}^{c_i}.
    \label{equ:tag_match}
\end{equation}
After bipartite matching of (\ref{equ:tag_match}), the pseudo-labels are $\{(\mathbf{\hat{b}}_{\hat{\sigma}(i)}, c_i)\}^G_{i=1}$, where
$\hat{\sigma}(i) \in \{1,...,K\}$ is the matched index to the $i$-th ground truth omni-label.
$\mathbf{\hat{b}}_{\hat{\sigma}(i)}$ is the predicted box and $c_i$ is the available ground truth class.

\vspace{-4mm}
\paragraph{TagsK} 
When the tags and their counts are known, $\mathbf{y}^o = \{(c_{j},n_{j})\}_{j=1}^{M}$, where $n_{j}$ is the number of objects of class $c_{j}$. There is no need to predict the counts anymore. The optimal matching can be computed with (\ref{equ:tag_match}), to obtain the pseudo-labels $\{(\mathbf{\hat{b}}_{\hat{\sigma}(i)}, c_i)\}^G_{i=1}$.

\vspace{-4mm}
\paragraph{PointsU}
When points of objects are known, $\mathbf{y}^o = \{\mathbf{g}_i\}=\{\mathbf{p}_i\in \mathcal{R}^2\}_{i=1}^{G}$, where $\mathbf{p}_i$ is a point and $G$ points in total. The matching cost is defined as
\vspace{-2mm}
\begin{equation}
    \mathcal{L}^p_{match}(\mathbf{g}_i,\mathbf{\hat{y}}_{\sigma(i)}) = (d_{i,\sigma(i)} + e_{i,\sigma(i)}) * \eta_{i,\sigma(i)},
    \label{equ:pointu_match}
\end{equation}
where $d_{i,\sigma(i)}$ is the $L$2 normalized distance, between the center of the predicted box and ground truth point, normalized to $[0,1]$ across $K \times G$ distances by min-max normalization, and $e_{i,\sigma(i)} = 1 \!-\! s_{\sigma(i)}$, where $s_{\sigma(i)}$ is the confidence score of $\sigma(i)$-th prediction. Finally, $\eta_{i,\sigma(i)}$ is an indicator: $\eta_{i,\sigma(i)}=1$ if the $i$-th ground truth point is inside the $\sigma(i)$-th predicted box, otherwise $+\infty$. This cost encourages the selected predicted box to cover the ground truth point with small geometric distance and high confidence. The pseudo-labels $\{(\mathbf{\hat{b}}_{\hat{\sigma}(i)}, \hat{c}_i)\}^G_{i=1}$ are obtained by optimizing (\ref{equ:pointu_match}) via Hungarian matching.

\vspace{-4mm}
\paragraph{PointsK}
When both the point and tag of an object are known, the ground truth is $\mathbf{y}^o = \{\mathbf{g}_i\} = \{(\mathbf{p}_i,c_{i})\in \mathcal{R}^2 \times \{1,2,...,C\}\}_{i=1}^{G}$. We combine (\ref{equ:tag_match}) and (\ref{equ:pointu_match}) linearly as the overall matching cost,
\begin{equation}
    \mathcal{L}_{match} (\mathbf{g}_i,\mathbf{\hat{y}}_{\sigma(i)}) = \gamma \mathcal{L}^t_{match} + (1-\gamma) \mathcal{L}^p_{match},
    \label{equ:point_gamma}
\end{equation}
where $\gamma$ is the trade-off coefficient, to obtain the pseudo-labels $\{(\mathbf{\hat{b}}_{\hat{\sigma}(i)}, c_i)\}^G_{i=1}$.

\vspace{-4mm}
\paragraph{Boxes}
When the bounding boxes are known but without classes, $\mathbf{y}^o = \{\mathbf{g}_i\} = \{\mathbf{b}_i \in \mathcal{R}^4\}^G_{i=1}$, we follow the bounding box cost definition of \cite{carion2020end}, 
\begin{equation}
    \mathcal{L}^b_{match}(\mathbf{g}_i,\mathbf{\hat{y}}_{\sigma(i)}) = \lambda_{\text{iou}} \mathcal{L}_{\text{iou}}(\mathbf{g}_i, \hat{\mathbf{b}}_{\sigma(i)}) + \lambda_{\text{L1}} ||\mathbf{g}_i - \hat{\mathbf{b}}_{\sigma(i)} ||_1,
    \label{equ:box_gamma}
\end{equation}
where $\mathcal{L}_{\text{iou}}$ is the generalized IoU loss~\cite{rezatofighi2019generalized}, to obtain the pseudo-labels $\{(\mathbf{b}_i, \hat{c}_{\hat{\sigma}(i)})\}^G_{i=1}$. Although BoxesEC and BoxesU have different box qualities, they are not differentiated by their matching costs here. 

The discussion above unifies pseudo-label filtering for all weak annotations as a bipartite matching problem, performed by global optimization on a set-to-set matching problem. This will be shown, in experiments, to outperform the heuristic choice of Section \ref{sec:simple_filter}.

\section{Experiments}

Omni-DETR is extensively evaluated on different datasets and settings.

\subsection{Experimental Settings}

\noindent\textbf{Datasets:} 
MS-COCO~\cite{lin2014microsoft}, PASCAL VOC~\cite{everingham2010pascal}, Bees~\cite{bees}, CrowdHuman~\cite{shao2018crowdhuman} and Objects365~\cite{shao2019objects365} are used for evaluation. To evaluate and compare Omni-DETR to methods addressing different problems, we use the multiple experimental settings of~\cite{liu2021unbiased,ren2020ufo,fang2021wssod}. 
(\uppercase\expandafter{\romannumeral1}) {\it COCO-standard}: we randomly sample $\{1, 2, 5, 10, 20, 30\} \%$ of data from COCO \emph{train2017} as the fully labeled training data and use the rest as the omni-labeled training data. 
(\uppercase\expandafter{\romannumeral2}) {\it COCO-35to80}: we use the COCO-35 (a.k.a. \emph{valminusminival}), a subset of 35K images of COCO \emph{train2017} as the fully labeled data and COCO-80, the COCO \emph{train2014} of 80K images, as the omni-labeled data. 
(\uppercase\expandafter{\romannumeral3}) {\it VOC-07to12}: we use the VOC07 \emph{trainval} as the fully labeled set and the VOC12 \emph{trainval} as the omni-labeled set.
On COCO, model performance is evaluated on the COCO \emph{val2017}, and VOC07 \emph{test} on VOC.\\
\noindent\textbf{Implementation details:} For fair comparison, ResNet-50 pretrained on ImageNet~\cite{deng2009imagenet,he2016deep} is used as the backbone. The confidence threshold $\tau=0.7$. For strong augmentation, following~\cite{zhu2020deformable,liu2021unbiased}, we apply random horizontal flipping, random resizing, random size cropping, color jittering, grayscale, Gaussian blur, and cutout patches. For weak augmentation, only random horizontal flipping is used. 
To mimic point annotations, we follow~\cite{chen2021points,ren2020ufo} and randomly sample a point from the instance mask if the dataset has instance segmentation, otherwise, we randomly sample a point inside each bounding box. 
For Extreme Clicking boxes, since \cite{papadopoulos2017extreme} only has partial annotations on VOC, we simulate similar annotations on other datasets by adding noise to their ground truth bounding box annotations, so that the resulting boxes have close distribution to that of Extreme Clicking on VOC. More details can be found in the supplementary.
The detection performance is evaluated with the teacher model for all experiments. We use $AP_{50:95}$, denoted as mAP, as the evaluation metric unless otherwise noted. The minimum size of image height and width is set to 600-pixels for faster experiments, except in experiments involving comparisons with other methods that use the standard 800-pixel size.

\begin{table}
\setlength{\abovecaptionskip}{-4.0pt}
\small
\begin{center}
\begin{tabular}{l|ccc}
 & mAP & AP$_{50}$ & AP$_{75}$\\
\thickhline
$10\%$ supervision &  28.0 &44.3 &29.5 \\
\quad + $90\%$ None &  32.4 &49.3 &34.5 \\
\quad+ $90\%$ TagsU &  34.7 & 52.4& 37.2\\
\quad+ $90\%$ TagsK &  35.2 &53.5 & 37.7\\
\quad+ $90\%$ PointsU & 34.1 &51.9 &36.2 \\
\quad+ $90\%$ PointsK & 35.7  &54.2 & 38.6\\
\quad+ $90\%$ BoxesEC &  36.4 & 54.6&39.3 \\
\quad+ $90\%$ BoxesU & 36.8  &54.8 &39.4
\end{tabular}
\end{center}
\caption{The effects of different weak annotations on the baseline of $10\%$ {\it COCO-standard} fully labeled data.}
\label{tab:single}
\end{table}

\subsection{Evaluation on Single Annotation}

Under the setting of {\it COCO-standard-$10\%$}, we first evaluate Omni-DETR for individual weak annotations in Table \ref{tab:single}, to study the effect of each weak annotation. The baseline is the standard supervised learning on $10\%$ labeled data. A few observations are available. First, the additional 90\% unlabeled data improves the baseline by 4.4\% when using semi-supervised learning. Annotating extra weak labels always enhances the performances by $1.7-4.4\%$. Second, among all annotation formats, PointsU has the smallest benefit and BoxesU the largest. Third, Extreme Clicking boxes (BoxesEC) is economical: only $0.3\%$ worse than the high-quality boxes of BoxesU but 5 times less costly. Fourth, count annotation provides a gain of $0.5\%$ over tag annotation (TagsU v.s. TagsK). Fifth, adding tag information to points (PointsU v.s. PointsK), leads to $1.6\%$ improvement.

\subsection{Comparison with the State-of-the-art}

Omni-DETR is compared with previous works under different settings. In this section, ``Supervised'' is the supervised Deformable DETR baseline trained on the available fully-labeled data only.

\begin{table}[t]
\setlength{\tabcolsep}{5pt}
\begin{center}
\small
\begin{tabular}{l|ccccc}
 & $1\%$ & $2\%$ &$5\%$ & $10\%$ & VOC\\
\thickhline
Faster R-CNN~\cite{ren2015faster,liu2021unbiased} &  9.1 & 12.7 & 18.5 &23.9 &42.1\\
Faster R-CNN$^*$ & 11.7 & 14.9 &  20.7 & 25.6 &42.6 \\
Deformable DETR$^*$& 11.0  &14.7 &  23.7 & 29.2 &46.2 \\
STAC~\cite{sohn2020simple} & 14.0 & 18.3 &  24.4 &28.6 & 44.6\\
Unbiased Teacher~\cite{liu2021unbiased} & {\bf20.8 } &{\bf 24.3} & 28.3  &31.5 & 48.7\\
Humble Teacher~\cite{Tang_2021_CVPR} &  17.0 &21.7 &27.7 &31.6 & 53.0\\
\hline
Omni-DETR  & 18.6  & 23.2 & {\bf 30.2}  &{\bf 34.1}& {\bf 53.4}
\end{tabular}
\caption{SSOD result comparison on {\it COCO-standard} and VOC-07to12. $^*$ indicates our implementation.}\vspace{-3mm}
\label{tab:semi}
\end{center}
\end{table}

\begin{table}[t]
\setlength{\tabcolsep}{6pt}
\begin{center}
\small
\begin{tabular}{l|ccc}
\centering
 & Supervised & +TagsU & +PointsK\\
\thickhline
UFO$^2$~\cite{ren2020ufo} & 29.1& 29.4 (+0.3)&30.1 (+1.0) \\
Omni-DETR & 34.3 & {\bf 39.4 (+5.1)} &{\bf 40.2 (+5.9)}\\
\end{tabular}
\caption{WSSOD comparison with UFO$^2$ on {\it COCO-35to80}. The numbers in parentheses are gains over the supervised baseline (Faster R-CNN for UFO$^2$ but Deformable DETR for ours).}\vspace{-3mm}
\label{tab:ufo}
\end{center}
\end{table}

\noindent\textbf{SSOD} When no annotation is available, the Omni-DETR becomes
a standard semi-supervised detector, which is compared to other SSOD methods in Table \ref{tab:semi}. We implemented the supervised Faster R-CNN and Deformable DETR trained only on the labeled data as the baselines.
Omni-DETR achieves the best results on $5\%$ and $10\%$ of COCO, and VOC, and comparable results with the state-of-the-art on $1\%$ and $2\%$ of COCO\footnote{\cite{li2021rethinking} mentioned Unbiased Teacher is weaker for smaller batch size.}. Note that our Omni-DETR is not designed specifically for SSOD, but it still achieves competitive results.

\noindent\textbf{WSSOD with tags} When additional tag annotation is available, SSOD becomes WSSOD with tags. We compare with the state-of-the-art methods, UFO$^2$~\cite{ren2020ufo} and Fang \etal~\cite{fang2021wssod}, on their settings. The results are reported in Table \ref{tab:ufo} and \ref{tab:wwsod_tags}, showing that our model consistently outperforms \cite{fang2021wssod,ren2020ufo}. It is worth noting, in Table \ref{tab:wwsod_tags}, that our model trained on $5\%$ ($10\%$) labeled data achieves $31.7$ ($35.9$) mAP, which is higher than \cite{fang2021wssod} trained on $10\%$ ($20\%$) labeled data. In Table \ref{tab:ufo}, we improve the supervised baseline by $5.1\%$ by using tags, whereas the improvement is $0.3\%$ for UFO$^2$. Our absolute improvement over UFO$^2$ is $10\%$.

\begin{table}
\setlength{\abovecaptionskip}{-1.0pt}
\setlength{\tabcolsep}{8pt}
\small
\begin{center}
\begin{tabular}{l|cccc}
 & $1\%$ &$5\%$ & $10\%$ & $20\%$\\
\thickhline
Supervised & 11.0   &  23.7 & 29.2 & 33.6 \\
Fang \etal~\cite{fang2021wssod} &18.4 & 27.4& 31.3&35.0\\
\hline
Omni-DETR (ours) &{\bf 20.1 } & {\bf 31.7}& {\bf 35.9}& {\bf 38.1}
\end{tabular}
\end{center}
\caption{WSSOD with tags result comparison on {\it COCO-standard}.}
\label{tab:wwsod_tags}
\end{table}

\noindent\textbf{WSSOD with points}
When additional point with tag annotation is available for SSOD, the problem becomes WSSOD with points~\cite{chen2021points}. Omni-DETR is compared with Point DETR and UFO$^2$, in Table \ref{tab:wwsod_points} and \ref{tab:ufo} respectively. It can be observed in Table \ref{tab:wwsod_points} that we outperform Point DETR by a significant margin ($5-7\%$). In Table \ref{tab:ufo}, when using points, Omni-DETR improves over the supervised baseline by $5.9\%$ where UFO$^2$ improves by $1\%$, and our absolute gain over UFO$^2$ is $10.1\%$.

\begin{table}
\setlength{\abovecaptionskip}{-2.0pt}
\setlength{\tabcolsep}{8pt}
\small
\begin{center}
\begin{tabular}{l|cccc}
 & $5\%$ &$10\%$ & $20\%$ & $30\%$\\
\thickhline
Supervised  & 23.7 & 29.2 & 33.6 & 35.2\\
Point DETR~\cite{chen2021points} &26.2 &30.4 &33.3 &34.8\\
\hline
Omni-DETR (ours) & {\bf 32.5} & {\bf 37.1}&{\bf 39.0} &{\bf 40.1}
\end{tabular}
\end{center}
\caption{WSSOD with points comparison on {\it COCO-standard}.}
\label{tab:wwsod_points}
\end{table}

\begin{table}[t]
\setlength{\tabcolsep}{8pt}
\begin{center}
\small
\begin{tabular}{l|ccc}
\centering
 & $80\%$B & $50\%$B & $20\%$B\\
\thickhline
UFO$^2$~\cite{ren2020ufo} & 14.1 & 11.1 &4.5 \\
Omni-DETR &{\bf 21.5}& {\bf 19.5} &{\bf 9.1}\\
\end{tabular}
\caption{OSOD result comparison with UFO$^2$ on COCO.}\vspace{-3mm}
\label{tab:ufo2}
\end{center}
\end{table}

\begin{table}
\setlength{\abovecaptionskip}{-2.0pt}
\setlength{\tabcolsep}{2pt}
\small
\begin{center}
\begin{tabular}{cccc|cccc}
    \multicolumn{4}{c|}{Simple Filtering}&\multicolumn{4}{c}{Unified Filtering}\\
\thickhline
    TagsU&TagsK&PointsU&PoinsK&TagsU&TagsK&PointsU&PointsK\\
    33.3&33.8& 32.4&34.6&34.7&35.2&34.1&35.7\\
\end{tabular}
\end{center}
\caption{Comparison with simple filters on {\it COCO-standard-$10\%$}.}
\label{tab:heuristic}
\end{table}

\begin{table}
\setlength{\abovecaptionskip}{-2.0pt}
\small
\begin{center}
\begin{tabular}{l|ccccc}
 & 0.5 &0.6 & 0.7 & 0.8 & 0.9\\
\thickhline
None & 28.9   &  31.5 & 32.4 & 31.4 &29.9\\
TagsU&31.1& 34.1& 34.7&33.9&33.1
\end{tabular}
\end{center}
\caption{Effect of $\tau$ on {\it COCO-standard-$10\%$}}
\label{tab:abla:t}
\end{table}

\begin{table}
\setlength{\abovecaptionskip}{-2.0pt}
\small
\begin{center}
\begin{tabular}{l|ccccc}
 & 0.00 &0.25 & 0.5 & 0.75 & 1.00\\
\thickhline
PointsK & 34.1   &  35.3 & 35.7 & 35.5 &35.2
\end{tabular}
\end{center}
\caption{Effect of $\gamma$ on {\it COCO-standard-$10\%$}.}
\label{tab:abla:gamma}
\end{table}

\noindent\textbf{OSOD} In addition to Table \ref{tab:ufo}, we also compare with UFO$^2$ on the X$\%B$ settings of \cite{ren2020ufo}, where X$\%B$ are different annotation policies using $10K$ images of COCO. Under a fixed budget, $X\%$ budget is spent on fully labeled annotations, and the rest on PointsK. As shown in Table \ref{tab:ufo2}, Omni-DETR still has consistently significant gains over UFO$^2$.

\subsection{Ablation Study}

We ablate some key components of our Omni-DETR on {\it COCO-standard-$10\%$} setting.

\noindent{\bf Comparison with simple filters} We compare the proposed unified pseudo-label filter with the simple filter of Section \ref{sec:simple_filter}.
Table \ref{tab:heuristic} shows that the proposed unified filter is better than the simple and heuristic filter under various settings. This is because the matching in the unified filter is a global solution by Hungarian algorithm, instead of a heuristic one as in the simple filter.

\begin{figure*}
  \centering
  \setlength{\tabcolsep}{2pt}
  \begin{tabular}{ccccc}
    \includegraphics[width=0.17\linewidth]{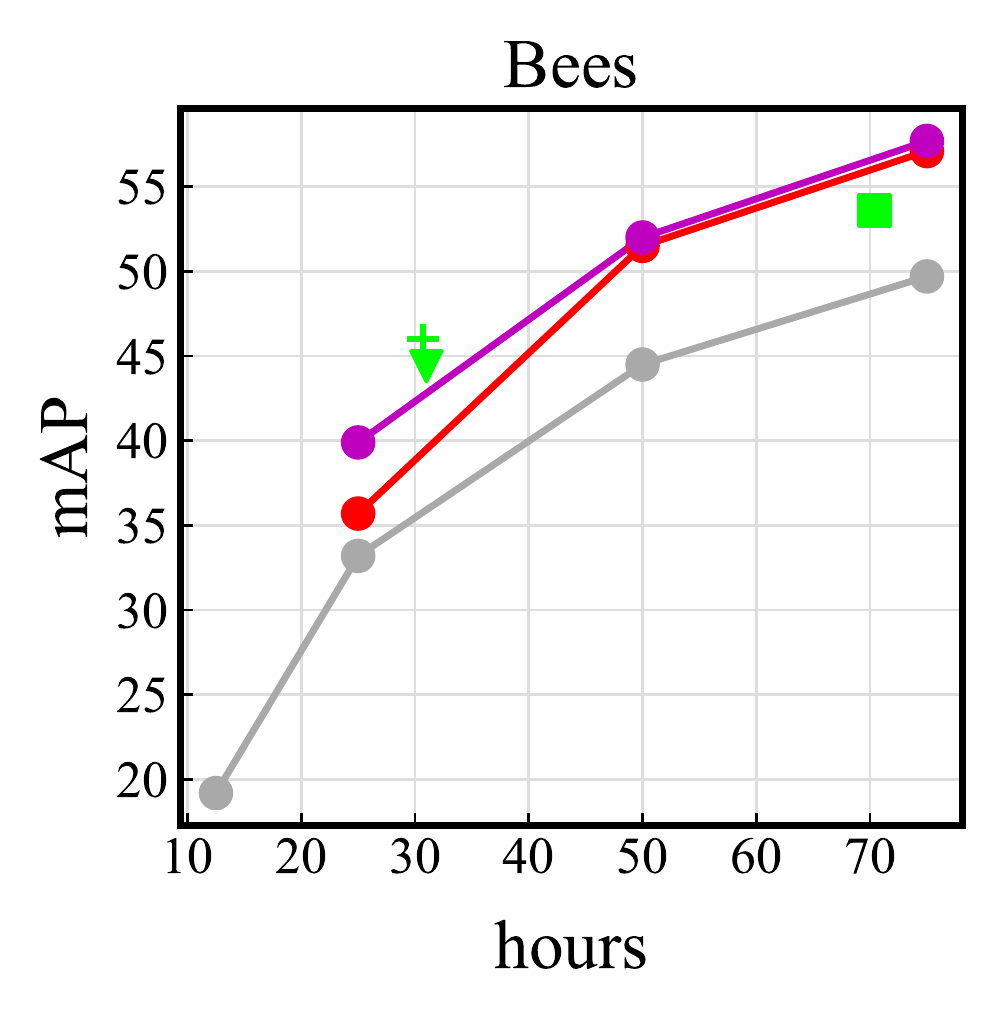}&
    \includegraphics[width=0.17\linewidth]{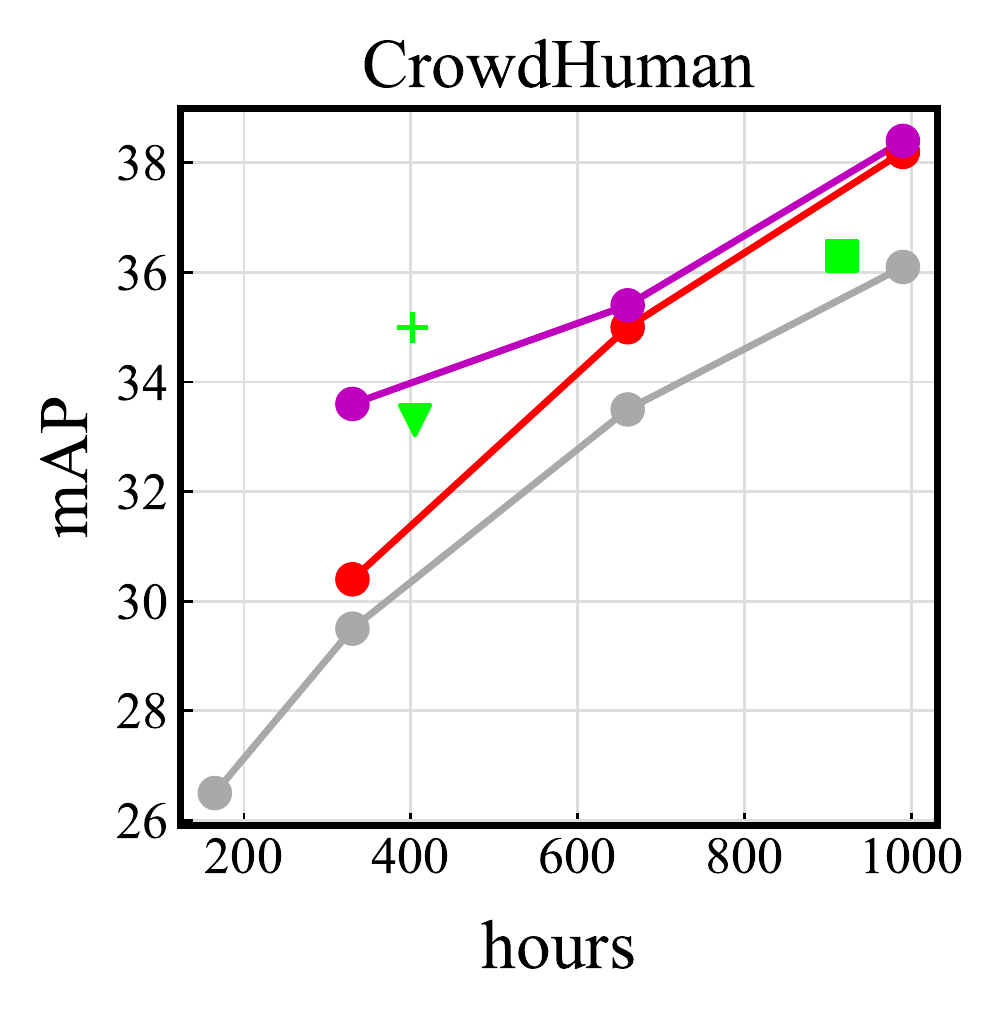} &
    \includegraphics[width=0.17\linewidth]{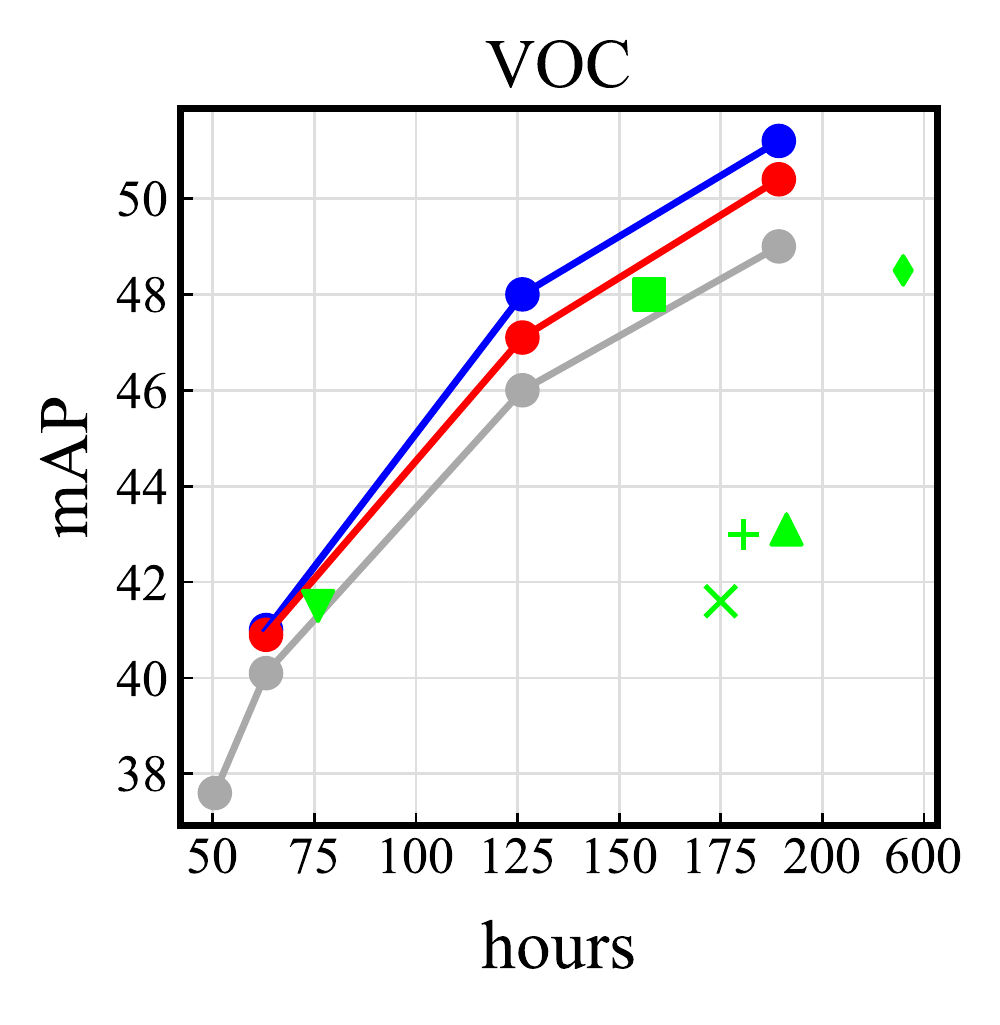} &
    \includegraphics[width=0.17\linewidth]{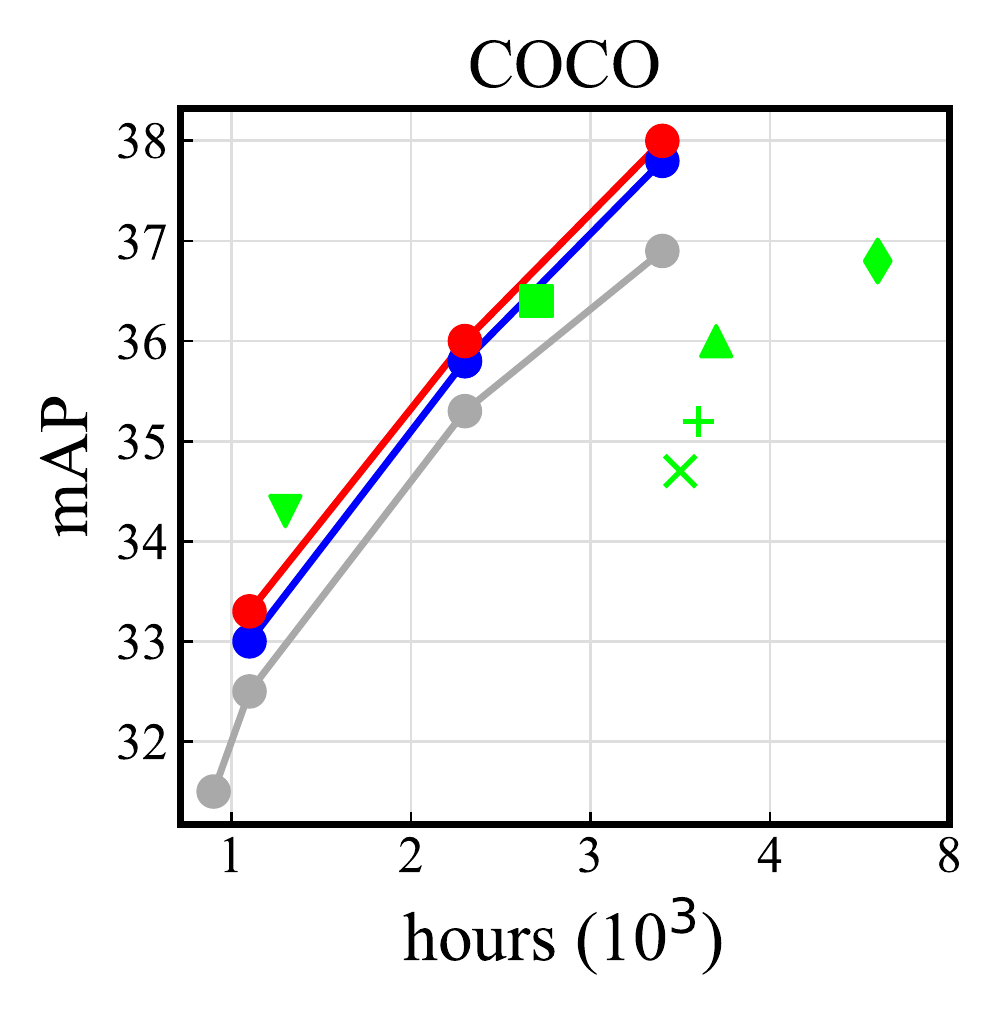} &
    \includegraphics[width=0.29\linewidth]{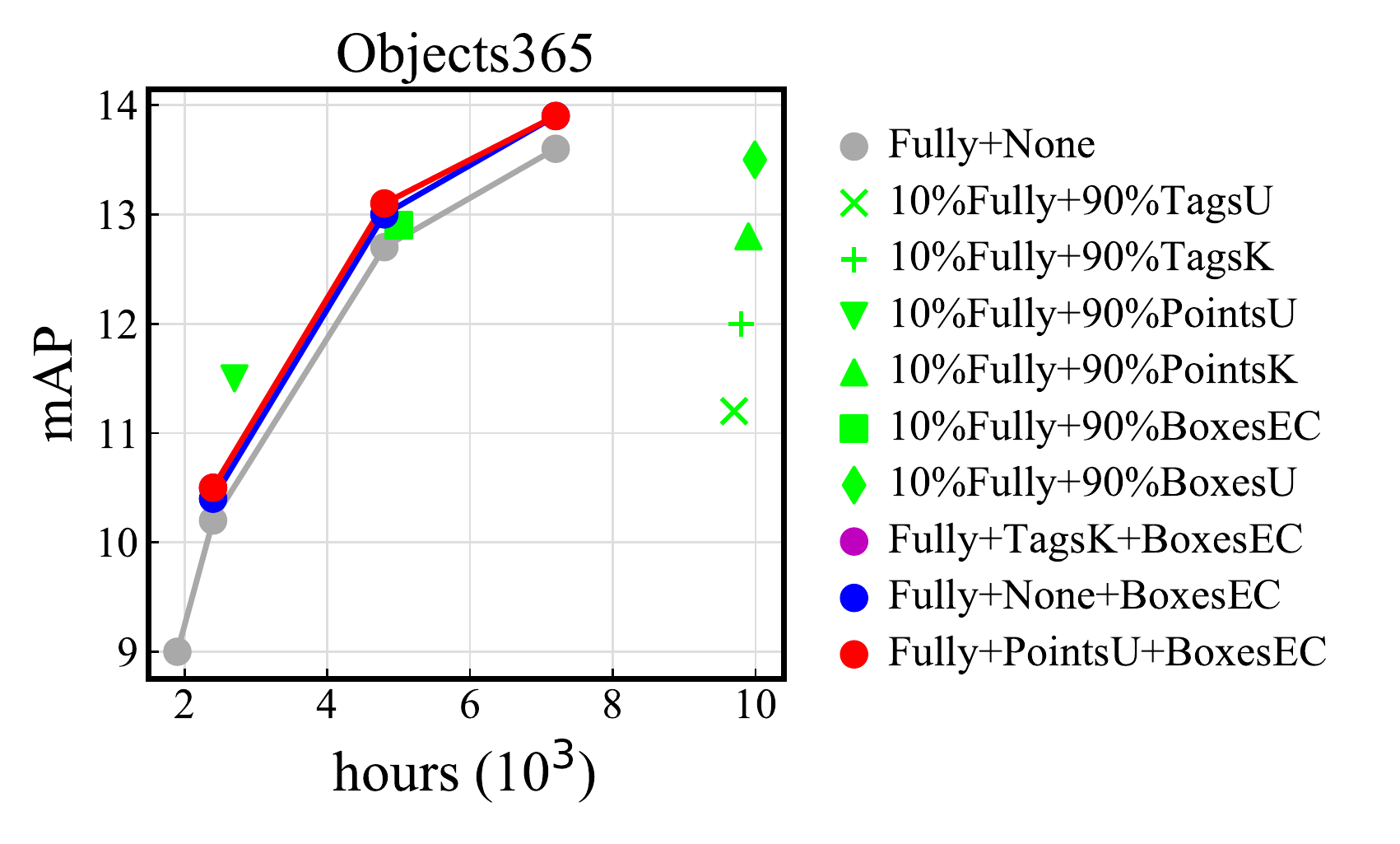}
  \end{tabular}
  \caption{\textbf{Accuracy (mAP) and annotation cost trade-off}. Grey lines are the SSOD baseline references. Green dots represent the WSSOD results with different weak labels. Red, blue and purple lines are the OSOD results for different mixture annotation choices.}
  \label{fig:map_cost}
\end{figure*}  

\noindent{\bf Confidence threshold} The confidence threshold $\tau$ used in Section \ref{sec:none_anno} and (\ref{equ:tagsU}) determines the trade-off between the quality and quantity of the pseudo-labels. A larger $\tau$ leads to fewer examples passing the threshold but with high quality, but a smaller $\tau$ allows more examples passing but more likely false positives. The results of different values of $\tau$ ($0.5$ to $0.9$) are reported in Table \ref{tab:abla:t}. $\tau=0.7$ is the best.

\noindent{\bf The effect of $\gamma$} The hyperparameter $\gamma$ of (\ref{equ:point_gamma}) balances the importance of positions and tag labels during the matching for point annotation. Its effect is evaluated in Table \ref{tab:abla:gamma}, and $\gamma=0.5$ is the best.

\noindent{\bf Pseudo bounding box} In Unbiased Teacher~\cite{liu2021unbiased}, pseudo bounding boxes are not used for learning from unlabeled data since the class confidence score does not reflect the goodness of the bounding box. However, we found that pseudo bounding boxes are useful and provide consistent improvement of $0.5-1\%$ in our experiments. One possible reason for the improvement is the higher quality of Omni-DETR pseudo bounding boxes.

\subsection{Budget-Aware Omni-Supervised Detection}
\label{subsec:budge-aware detection}

We also empirically study the trade-off between annotation cost\footnote{Only human annotation costs are considered, and other costs are ignored if there is any.} and accuracy of several annotation policies. 
Here, annotation policy refers to the strategy for mixing different annotation formats. 
Five diverse datasets with different characteristics are tested. The annotation cost, in seconds per image, for each type of annotation is shown in Table \ref{tab:cost},  following~\cite{ren2020ufo,bearman2016s,su2012crowdsourcing,papadopoulos2017extreme}. For each dataset, we attempt to identify the best annotation policy, given different budgets. SSOD is used as the baseline because, when the entire budget is used on standard full object annotation (bounding boxes and tags), the remaining data is considered unlabeled, which is the standard SSOD setup and widely adopted in practice. Next, we consider different weakly semi-supervised settings with $10\%$ data fully labeled and the remaining $90\%$ labeled with different weak annotations. Finally, two choices of mixture annotation are tested under three budgets per dataset, to show the omni-supervised results. We only tested
the combination of weak annotations that are better than the SSOD baseline, and the combination ratio was decided manually so that the mixture and full annotations had similar costs, for a fair comparison. Please see more details in the supplementary.

The results are summarized in Figure \ref{fig:map_cost}. The grey lines and green dots are the SSOD and WSSOD results, respectively, whereas red/blue/purple lines are for OSOD variants. It can be found that the OSOD results are higher than the strong SSOD baseline in general. For a target accuracy, the mixture annotation strategy can help reduce cost, and for a given cost, SSOD can improve accuracy. For example, on Bees at the accuracy of $40\%$ mAP, using mixture annotations of TagsK and PointsU can save the cost of about 15 hours from the standard detection annotation (25 hours v.s. 40 hours). On CrowdHuman, for the cost of about 330 hours OSOD improves the strong SSOD baseline mAP by $\sim$4\%. These findings support our claim that weak annotations are useful and can achieve a better cost-accuracy trade-off than standard detection annotation.

Some additional interesting observations are also available in Figure \ref{fig:map_cost}. First, the green upside-down triangles, rectangles and red/blue lines are all higher than the strong SSOD baseline. This suggests that annotating points (PointsU) and/or Extreme Clicking boxes (BoxesEC) is a better choice than standard complete annotation. Second, the green plus is above the reference on Bees and CrowdHuman, but not for the other three datasets,  indicating that count annotation (TagsK) is useful for datasets with dense objects. Finally, weak annotations such as TagsU (green cross), TagsK (green plus) and PointsK (green triangle), are far below the SSOD baseline on datasets like VOC, COCO and Objects365. This suggests that tags are not a good annotation format for datasets with large number of classes, where annotating tags is expensive. In general, the optimal annotation choice is quite dataset-specific, depending on characteristics such as number of categories, number of objects per image, object size, etc. 

\newcommand{\rrr}[1]{\rotatebox{60}{#1}}
\begin{table}
\setlength{\abovecaptionskip}{-2.0pt}
\setlength{\tabcolsep}{3pt}
\small
\begin{center}
\begin{tabular}{l|ccccccc}
Datasets&\rrr{TagsU} & \rrr{TagsK} & \rrr{PointsU} &\rrr{PointsK} & \rrr{BoxesEC} & \rrr{BoxesU} & \rrr{Fully}\\
\thickhline
Bees & -&6.1& 6.4& 6.4& 50 & 249.9&249.9\\
CrowdHuamn & -& 19.4& 20.4&20.4 &158.5 &792.4 & 792.4 \\
VOC &20 & 21& 2.2& 22.9& 16.8& 84&102.6\\
COCO & 80 &84.2 &6.9&88.7& 53.9 & 269.5&346\\
Objects365&365  & 375.8&14.2 &381.7 & 110.6&553 &913
\end{tabular}
\end{center}
\caption{Labeling cost estimation for different annotations (seconds per image).}
\label{tab:cost}
\end{table}

\vspace{-2mm}
\section{Conclusion}
We have proposed a unified framework for omni-supervised object detection, which can use different types of weak annotations. With this unified framework, we have found weak annotations are helpful and a mixture of them can achieve a better cost-accuracy trade-off.\\
\noindent\textbf{Limitations and Potential Negative Social Impact:} 
It is unclear whether these findings of this paper are still consistent on larger datasets, since we have not explored dataset size beyond COCO ($\sim$120K images) yet.
In addition, Omni-DETR could potentially increase the risk of improper use of detection systems, because it makes good detectors more accessible with fewer annotation efforts.

{\small
\bibliographystyle{ieee_fullname}
\bibliography{egbib}

\begin{thebibliography}{10}\itemsep=-1pt

\bibitem{DBLP:conf/iclr/AthiwaratkunFIW19}
Ben Athiwaratkun, Marc Finzi, Pavel Izmailov, and Andrew~Gordon Wilson.
\newblock There are many consistent explanations of unlabeled data: Why you
  should average.
\newblock In {\em ICLR}, 2019.

\bibitem{bearman2016s}
Amy Bearman, Olga Russakovsky, Vittorio Ferrari, and Li Fei-Fei.
\newblock What’s the point: Semantic segmentation with point supervision.
\newblock In {\em ECCV}, pages 549--565. Springer, 2016.

\bibitem{bees}
Bees.
\newblock \url{https://lila.science/datasets/boxes-on-bees-and-pollen}.

\bibitem{bilen2016weakly}
Hakan Bilen and Andrea Vedaldi.
\newblock Weakly supervised deep detection networks.
\newblock In {\em CVPR}, pages 2846--2854, 2016.

\bibitem{cai2021exponential}
Zhaowei Cai, Avinash Ravichandran, Subhransu Maji, Charless Fowlkes, Zhuowen
  Tu, and Stefano Soatto.
\newblock Exponential moving average normalization for self-supervised and
  semi-supervised learning.
\newblock In {\em CVPR}, pages 194--203, 2021.

\bibitem{cai2018cascade}
Zhaowei Cai and Nuno Vasconcelos.
\newblock Cascade r-cnn: Delving into high quality object detection.
\newblock In {\em CVPR}, pages 6154--6162, 2018.

\bibitem{carion2020end}
Nicolas Carion, Francisco Massa, Gabriel Synnaeve, Nicolas Usunier, Alexander
  Kirillov, and Sergey Zagoruyko.
\newblock End-to-end object detection with transformers.
\newblock In {\em ECCV}, pages 213--229. Springer, 2020.

\bibitem{chandra2020active}
Akshay~L Chandra, Sai~Vikas Desai, Vineeth~N Balasubramanian, Seishi Ninomiya,
  and Wei Guo.
\newblock Active learning with point supervision for cost-effective panicle
  detection in cereal crops.
\newblock {\em Plant Methods}, 16(1):1--16, 2020.

\bibitem{chen2021points}
Liangyu Chen, Tong Yang, Xiangyu Zhang, Wei Zhang, and Jian Sun.
\newblock Points as queries: Weakly semi-supervised object detection by points.
\newblock In {\em CVPR}, pages 8823--8832, 2021.

\bibitem{counting}
Counting.
\newblock
  \url{http://www.blog.republicofmath.com/how-long-does-it-take-to-count-to-one-trillion/}.

\bibitem{deng2009imagenet}
Jia Deng, Wei Dong, Richard Socher, Li-Jia Li, Kai Li, and Li Fei-Fei.
\newblock Imagenet: A large-scale hierarchical image database.
\newblock In {\em CVPR}, pages 248--255. Ieee, 2009.

\bibitem{dollar2009pedestrian}
Piotr Doll{\'a}r, Christian Wojek, Bernt Schiele, and Pietro Perona.
\newblock Pedestrian detection: A benchmark.
\newblock In {\em CVPR}, pages 304--311. IEEE, 2009.

\bibitem{everingham2010pascal}
Mark Everingham, Luc Van~Gool, Christopher~KI Williams, John Winn, and Andrew
  Zisserman.
\newblock The pascal visual object classes (voc) challenge.
\newblock {\em IJCV}, 88(2):303--338, 2010.

\bibitem{fang2021wssod}
Shijie Fang, Yuhang Cao, Xinjiang Wang, Kai Chen, Dahua Lin, and Wayne Zhang.
\newblock Wssod: A new pipeline for weakly-and semi-supervised object
  detection.
\newblock {\em arXiv preprint arXiv:2105.11293}, 2021.

\bibitem{gao2019note}
Jiyang Gao, Jiang Wang, Shengyang Dai, Li-Jia Li, and Ram Nevatia.
\newblock Note-rcnn: Noise tolerant ensemble rcnn for semi-supervised object
  detection.
\newblock In {\em ICCV}, pages 9508--9517, 2019.

\bibitem{girshick2015fast}
Ross Girshick.
\newblock Fast r-cnn.
\newblock In {\em ICCV}, pages 1440--1448, 2015.

\bibitem{girshick2014rich}
Ross Girshick, Jeff Donahue, Trevor Darrell, and Jitendra Malik.
\newblock Rich feature hierarchies for accurate object detection and semantic
  segmentation.
\newblock In {\em CVPR}, pages 580--587, 2014.

\bibitem{gokberk2014multi}
Ramazan Gokberk~Cinbis, Jakob Verbeek, and Cordelia Schmid.
\newblock Multi-fold mil training for weakly supervised object localization.
\newblock In {\em CVPR}, pages 2409--2416, 2014.

\bibitem{gupta2019lvis}
Agrim Gupta, Piotr Dollar, and Ross Girshick.
\newblock Lvis: A dataset for large vocabulary instance segmentation.
\newblock In {\em CVPR}, pages 5356--5364, 2019.

\bibitem{gygli2019efficient}
Michael Gygli and Vittorio Ferrari.
\newblock Efficient object annotation via speaking and pointing.
\newblock {\em IJCV}, pages 1--15, 2019.

\bibitem{he2016deep}
Kaiming He, Xiangyu Zhang, Shaoqing Ren, and Jian Sun.
\newblock Deep residual learning for image recognition.
\newblock In {\em CVPR}, pages 770--778, 2016.

\bibitem{DBLP:conf/uai/IzmailovPGVW18}
Pavel Izmailov, Dmitrii Podoprikhin, Timur Garipov, Dmitry~P. Vetrov, and
  Andrew~Gordon Wilson.
\newblock Averaging weights leads to wider optima and better generalization.
\newblock In {\em UAI}, pages 876--885. {AUAI} Press, 2018.

\bibitem{jeong2019consistency}
Jisoo Jeong, Seungeui Lee, Jeesoo Kim, and Nojun Kwak.
\newblock Consistency-based semi-supervised learning for object detection.
\newblock {\em NeurIPS}, 32:10759--10768, 2019.

\bibitem{jie2017deep}
Zequn Jie, Yunchao Wei, Xiaojie Jin, Jiashi Feng, and Wei Liu.
\newblock Deep self-taught learning for weakly supervised object localization.
\newblock In {\em CVPR}, pages 1377--1385, 2017.

\bibitem{kuhn1955hungarian}
Harold~W Kuhn.
\newblock The hungarian method for the assignment problem.
\newblock {\em Naval research logistics quarterly}, 2(1-2):83--97, 1955.

\bibitem{kuznetsova2018open}
Alina Kuznetsova, Hassan Rom, Neil Alldrin, Jasper Uijlings, Ivan Krasin, Jordi
  Pont-Tuset, Shahab Kamali, Stefan Popov, Matteo Malloci, Alexander
  Kolesnikov, et~al.
\newblock The open images dataset v4: Unified image classification, object
  detection, and visual relationship detection at scale.
\newblock {\em arXiv preprint arXiv:1811.00982}, 2018.

\bibitem{li2021rethinking}
Hengduo Li and et al.
\newblock Rethinking pseudo labels for semi-supervised object detection.
\newblock {\em arXiv:2106.00168}, 2021.

\bibitem{lin2017feature}
Tsung-Yi Lin, Piotr Doll{\'a}r, Ross Girshick, Kaiming He, Bharath Hariharan,
  and Serge Belongie.
\newblock Feature pyramid networks for object detection.
\newblock In {\em CVPR}, pages 2117--2125, 2017.

\bibitem{lin2014microsoft}
Tsung-Yi Lin, Michael Maire, Serge Belongie, James Hays, Pietro Perona, Deva
  Ramanan, Piotr Doll{\'a}r, and C~Lawrence Zitnick.
\newblock Microsoft coco: Common objects in context.
\newblock In {\em ECCV}, pages 740--755. Springer, 2014.

\bibitem{liu2016ssd}
Wei Liu, Dragomir Anguelov, Dumitru Erhan, Christian Szegedy, Scott Reed,
  Cheng-Yang Fu, and Alexander~C Berg.
\newblock Ssd: Single shot multibox detector.
\newblock In {\em ECCV}, pages 21--37. Springer, 2016.

\bibitem{liu2021unbiased}
Yen-Cheng Liu, Chih-Yao Ma, Zijian He, Chia-Wen Kuo, Kan Chen, Peizhao Zhang,
  Bichen Wu, Zsolt Kira, and Peter Vajda.
\newblock Unbiased teacher for semi-supervised object detection.
\newblock {\em arXiv preprint arXiv:2102.09480}, 2021.

\bibitem{papadopoulos2017extreme}
Dim~P Papadopoulos, Jasper~RR Uijlings, Frank Keller, and Vittorio Ferrari.
\newblock Extreme clicking for efficient object annotation.
\newblock In {\em ICCV}, pages 4930--4939, 2017.

\bibitem{papadopoulos2017training}
Dim~P Papadopoulos, Jasper~RR Uijlings, Frank Keller, and Vittorio Ferrari.
\newblock Training object class detectors with click supervision.
\newblock In {\em CVPR}, pages 6374--6383, 2017.

\bibitem{redmon2016you}
Joseph Redmon, Santosh Divvala, Ross Girshick, and Ali Farhadi.
\newblock You only look once: Unified, real-time object detection.
\newblock In {\em CVPR}, pages 779--788, 2016.

\bibitem{ren2015faster}
Shaoqing Ren, Kaiming He, Ross Girshick, and Jian Sun.
\newblock Faster r-cnn: Towards real-time object detection with region proposal
  networks.
\newblock {\em NeurIPS}, 28:91--99, 2015.

\bibitem{ren2020ufo}
Zhongzheng Ren, Zhiding Yu, Xiaodong Yang, Ming-Yu Liu, Alexander~G Schwing,
  and Jan Kautz.
\newblock Ufo$^2$: A unified framework towards omni-supervised object
  detection.
\newblock In {\em ECCV}, pages 288--313. Springer, 2020.

\bibitem{rezatofighi2019generalized}
Hamid Rezatofighi, Nathan Tsoi, JunYoung Gwak, Amir Sadeghian, Ian Reid, and
  Silvio Savarese.
\newblock Generalized intersection over union: A metric and a loss for bounding
  box regression.
\newblock In {\em CVPR}, pages 658--666, 2019.

\bibitem{rosenberg2005semi}
Chuck Rosenberg, Martial Hebert, and Henry Schneiderman.
\newblock Semi-supervised self-training of object detection models.
\newblock 2005.

\bibitem{shao2019objects365}
Shuai Shao, Zeming Li, Tianyuan Zhang, Chao Peng, Gang Yu, Xiangyu Zhang, Jing
  Li, and Jian Sun.
\newblock Objects365: A large-scale, high-quality dataset for object detection.
\newblock In {\em CVPR}, pages 8430--8439, 2019.

\bibitem{shao2018crowdhuman}
Shuai Shao, Zijian Zhao, Boxun Li, Tete Xiao, Gang Yu, Xiangyu Zhang, and Jian
  Sun.
\newblock Crowdhuman: A benchmark for detecting human in a crowd.
\newblock {\em arXiv preprint arXiv:1805.00123}, 2018.

\bibitem{sohn2020fixmatch}
Kihyuk Sohn, David Berthelot, Chun-Liang Li, Zizhao Zhang, Nicholas Carlini,
  Ekin~D Cubuk, Alex Kurakin, Han Zhang, and Colin Raffel.
\newblock Fixmatch: Simplifying semi-supervised learning with consistency and
  confidence.
\newblock In {\em NeurIPS}, 2020.

\bibitem{sohn2020simple}
Kihyuk Sohn, Zizhao Zhang, Chun-Liang Li, Han Zhang, Chen-Yu Lee, and Tomas
  Pfister.
\newblock A simple semi-supervised learning framework for object detection.
\newblock {\em arXiv preprint arXiv:2005.04757}, 2020.

\bibitem{song2014weakly}
Hyun~Oh Song, Yong~Jae Lee, Stefanie Jegelka, and Trevor Darrell.
\newblock Weakly-supervised discovery of visual pattern configurations.
\newblock {\em arXiv preprint arXiv:1406.6507}, 2014.

\bibitem{su2012crowdsourcing}
Hao Su, Jia Deng, and Li Fei-Fei.
\newblock Crowdsourcing annotations for visual object detection.
\newblock In {\em AAAI workshop}, 2012.

\bibitem{tang2018pcl}
Peng Tang, Xinggang Wang, Song Bai, Wei Shen, Xiang Bai, Wenyu Liu, and Alan
  Yuille.
\newblock Pcl: Proposal cluster learning for weakly supervised object
  detection.
\newblock {\em TPAMI}, 42(1):176--191, 2018.

\bibitem{Tang_2021_CVPR}
Yihe Tang, Weifeng Chen, Yijun Luo, and Yuting Zhang.
\newblock Humble teachers teach better students for semi-supervised object
  detection.
\newblock In {\em CVPR}, pages 3132--3141, June 2021.

\bibitem{tarvainen2017mean}
Antti Tarvainen and Harri Valpola.
\newblock Mean teachers are better role models: Weight-averaged consistency
  targets improve semi-supervised deep learning results.
\newblock {\em arXiv preprint arXiv:1703.01780}, 2017.

\bibitem{tian2019fcos}
Zhi Tian, Chunhua Shen, Hao Chen, and Tong He.
\newblock Fcos: Fully convolutional one-stage object detection.
\newblock In {\em ICCV}, pages 9627--9636, 2019.

\bibitem{vaswani2017attention}
Ashish Vaswani, Noam Shazeer, Niki Parmar, Jakob Uszkoreit, Llion Jones,
  Aidan~N Gomez, {\L}ukasz Kaiser, and Illia Polosukhin.
\newblock Attention is all you need.
\newblock In {\em NeurIPS}, pages 5998--6008, 2017.

\bibitem{xie2020self}
Qizhe Xie, Minh-Thang Luong, Eduard Hovy, and Quoc~V Le.
\newblock Self-training with noisy student improves imagenet classification.
\newblock In {\em CVPR}, pages 10687--10698, 2020.

\bibitem{DBLP:conf/cvpr/YangWW0021}
Qize Yang, Xihan Wei, Biao Wang, Xian{-}Sheng Hua, and Lei Zhang.
\newblock Interactive self-training with mean teachers for semi-supervised
  object detection.
\newblock In {\em CVPR}, pages 5941--5950, 2021.

\bibitem{DBLP:conf/cvpr/0001YWQL21}
Qiang Zhou, Chaohui Yu, Zhibin Wang, Qi Qian, and Hao Li.
\newblock Instant-teaching: An end-to-end semi-supervised object detection
  framework.
\newblock In {\em CVPR}, pages 4081--4090, 2021.

\bibitem{zhu2020deformable}
Xizhou Zhu, Weijie Su, Lewei Lu, Bin Li, Xiaogang Wang, and Jifeng Dai.
\newblock Deformable detr: Deformable transformers for end-to-end object
  detection.
\newblock {\em arXiv preprint arXiv:2010.04159}, 2020.

\end{thebibliography}
}

\newpage
\appendix

\section{Experimental Implementation Details}

In this supplement, we show the details that are not presented in the main paper due to the page limitation.

\subsection{Annotation Cost Calculation in Table \ref{tab:cost}}

In this section, we explain how the numbers in Table \ref{tab:cost} are calculated. It is non-trivial to compute the labeling time for each type of annotation because it depends on several factors like the annotation tools or platforms, the quality requirement of the annotations, the crowdsourcing protocol used, etc. In our work, we mainly follow \cite{bearman2016s,ren2020ufo,papadopoulos2017extreme,su2012crowdsourcing,counting,lin2014microsoft} for the calculation. 

We denote the averaged number of categories per image as $C_{avg}$, the averaged number of instances per image as $I_{avg}$, and the overall number of categories for a dataset as $C$. We first list the statistics information for each dataset in Table \ref{tab:set:stat}. Then we consider the labeling time calculation for each weak annotation.

\vspace{-5mm}
\paragraph{TagsU} According to \cite{bearman2016s,ren2020ufo}, collecting image-level class labels takes $\sim{1}$ second per category per image. Thus, the expected annotation time is equal to $C$ on COCO, VOC, Objects365. On Bees or CrowdHuman, TagsU is not considered since they only have one category.

\vspace{-4mm}
\paragraph{PointsU} According to \cite{bearman2016s}, it takes 0.9 seconds on average to annotate one point. Thus, the time is $0.9 \times I_{avg}$.

\vspace{-4mm}
\paragraph{PointsK} We follow the computation of \cite{ren2020ufo}. It takes $1$ second to eliminate every non-existing
class, and $C - C_{avg}$ seconds in total. \cite{ren2020ufo} reports that annotators
take a median of 2.4 seconds to click on the first instance of a class and 0.9 seconds for every additional instance. Thus the total labeling time is $(C - C_{avg})+2.4\times C_{avg}+0.9 \times(I_{avg}-C_{avg})$ on COCO, VOC, Objects365. On Bees or CrowdHuman, the time is equal to that of PointsU because since they only have one category.

\vspace{-4mm}
\paragraph{TagsK} It takes about $1$ second to count a number~\cite{counting}. Thus, on COCO, VOC, Objects365, 
the estimated time is $(C - C_{avg})+1.0\times C_{avg}+1.0 \times(I_{avg}-C_{avg}) = C+I_{avg}-C_{avg}$.
Because this computation is derived from multi-class data domains~\cite{ren2020ufo}, it is not applicable on Bees or CrowdHuman where they have only one category. For this reason,
we simply estimate the TagsK cost of the single-class dataset as $k$ times of the PointsK cost. Here $k$ is the averaged proportion of TagsK cost over PointsK cost on three multi-class datasets (VOC, COCO and Objects365), i.e., $k = (21/22.9+84.2/88.7+375.8/381.7)/3 = 0.95$, where these numbers are from Table \ref{tab:cost} (the columns of TagsK and PointsK cost). Thus, the costs of TagsK on Bees and CrowdHuman are computed by $0.95 \times 6.4=6.1$, $0.95 \times 20.4 = 19.4$, respectively, where these numbers are from Table \ref{tab:cost} (the columns of TagsK and PointsK cost).

\vspace{-4mm}
\paragraph{BoxesEC} \cite{papadopoulos2017extreme} reports that it takes $7$ seconds for one Extreme Clicking box, so the time is $7 \times I_{avg}$ seconds.

\vspace{-4mm}
\paragraph{BoxesU} Similarly, since annotating a high quality box needs $35$ seconds~\cite{su2012crowdsourcing}, it takes $35 \times I_{avg}$ seconds per image.

\vspace{-4mm}
\paragraph{Fully} Following~\cite{ren2020ufo},
the total time for full annotation is computed by $(C - C_{avg}) + 35 \times I_{avg}$ on COCO, VOC, Objects365. On Bees and CrowdHuman, the time is equal to that of BoxesU because there is no category labeling.

\begin{table}
\setlength{\abovecaptionskip}{-1.0pt}
\setlength{\tabcolsep}{4pt}
\small
\begin{center}
\begin{tabular}{l|ccccc}
 & COCO &VOC & Objects365 & Bees & CrowdHuman\\
\thickhline
$C$ & 80   &  20 & 365 & 1 &1\\
$C_{avg}$ & 3.5   &  1.4 & 5 & 1 &1\\
$I_{avg}$ & 7.7   &  2.4 & 15.8 & 7.14 &22.64
\end{tabular}
\end{center}
\caption{Dataset statistic. The information is provided by \cite{everingham2010pascal,lin2014microsoft,su2012crowdsourcing,shao2019objects365,bees,ren2020ufo}.}
\label{tab:set:stat}
\end{table}

\begin{table*}
\setlength{\abovecaptionskip}{-1.0pt}
\setlength{\tabcolsep}{2pt}
\small
\begin{center}
\begin{tabular}{l|l|ccccc|cc}
Dataset & Omni-label & Fully ($\%$) & None ($\%$)& TagsK ($\%$) & PointsU ($\%$) & BoxesEC ($\%$) & cost (hours) & mAP\\
\thickhline
\multirow{6}{*}{Bees} & \multirow{3}{*}{Fully+TagsK+BoxesEC} & 5& 0& 80&0 &15 &25 &39.9 \\
&   & 10& 0 & 46 &0  &44&50 &52.0 \\ 
&   &20 & 0 & 34 & 0 &46& 75&57.7 \\ \cline{2-9}
 & \multirow{3}{*}{Fully+PointsU+BoxesEC} & 5& 0& 0& 80 & 15& 25&35.7 \\
&   &10 & 0 &0 &46 &44& 50&51.5 \\ 
&   &  20&0 &0 &34  &46& 75&57.1 \\ \cline{2-9}
\hline
\multirow{6}{*}{CrowdHuman} & \multirow{3}{*}{Fully+TagsK+BoxesEC} & 5& 0& 80& 0 &15 &330 &33.6 \\
&   & 10& 0 & 46 &0  &44&660 &35.4 \\ 
&   &20 & 0 & 34 & 0 &46&990 &38.4 \\ \cline{2-9}
 & \multirow{3}{*}{Fully+PointsU+BoxesEC} & 5& 0& 0& 80 & 15& 330&30.4 \\
&   &10 & 0 &0 &46 &44&660 &35.0\\ 
&   &  20&0 &0  &34  &46& 990&38.2 \\ \cline{2-9}
\hline
\multirow{6}{*}{VOC} & \multirow{3}{*}{Fully+None+BoxesEC} & 8&80 &0 &  0& 12& 63.1&41.0 \\
&   &10 & 29 & 0 & 0 &61& 126.2&48.0 \\ 
&   &  20&19 & 0 & 0 &61& 189.3&51.2 \\ \cline{2-9}
 & \multirow{3}{*}{Fully+PointsU+BoxesEC} & 8&0 &0 &  91& 1&63.1 &40.9 \\
 &   & 10 & 0& 0 & 33 &57& 126.2&47.1 \\ 
&   & 20 & 0&0 &22 &58&189.3 &50.4 \\ \cline{2-9}
\hline
\multirow{6}{*}{COCO} & \multirow{3}{*}{Fully+None+BoxesEC} &8 &79 & 0& 0 & 13& 1.1$\times 10^3$&33.0 \\
&   &  10 &26& 0 & 0 &64& 2.3$\times 10^3$&35.8 \\ 
&   & 20  & 16 &0& 0 &64& 3.4$\times 10^3$&37.8 \\ \cline{2-9}
 & \multirow{3}{*}{Fully+PointsU+BoxesEC} & 8&0 & 0& 91 &1&1.1 $\times 10^3$& 33.3 \\
 &   & 10&  0 & 0 & 30 &60& 2.3$\times 10^3$&36.0 \\ 
&   & 20 & 0& 0&18 &62& 3.4$\times 10^3$&38.0 \\ \cline{2-9}
\hline
\multirow{6}{*}{Objects365} & \multirow{3}{*}{Fully+None+BoxesEC} & 8& 75& 0& 0 &17 &2.4$\times 10^3$ &10.4\\
&   & 10& 7 & 0 &0  &83&4.8$\times 10^3$ &13.0 \\ 
&   & 25  &25 &0 & 0 &50& 7.2$\times 10^3$&13.9 \\ \cline{2-9}
 & \multirow{3}{*}{Fully+PointsU+BoxesEC} & 8&0 &0 & 86 & 6&2.4$\times 10^3$ &10.5\\
 &   & 10 &0 &0  &8  &82& 4.8$\times 10^3$&13.1 \\ 
&   & 25 &0 &0 &34 &41& 7.2$\times 10^3$&13.9 \\
\hline
\end{tabular}
\end{center}
\caption{The details of omni-supervision experiments in Section \ref{subsec:budge-aware detection}.}
\label{tab:omni-per}
\end{table*}

\subsection{Datasets and Splitting Details in Section \ref{subsec:budge-aware detection}}

In the paper, for each dataset, Figure \ref{fig:map_cost} shows two different mixture policies, and three different budgets for each mixture policy. Table \ref{tab:omni-per} reports the detailed mixture percentages and other information. The training set used to be split into labeled and omni-labeled data is presented as follows for each dataset.

\vspace{-4mm}
\paragraph{Bees~\cite{bees}} The total number of images is 3,596. Since Bees does not split the dataset officially, we randomly sample $80\%$ images as the training set, after removing the broken images. The model is evaluated on the rest $20\%$ data.

\vspace{-4mm}
\paragraph{CrowdHuman~\cite{shao2018crowdhuman}} The official training set of 15,000 images is split by different percentages for the omni-supervision experiments. The model is evaluated on the official validation set.

\vspace{-4mm}
\paragraph{VOC~\cite{everingham2010pascal}} We combine VOC07 \emph{trainval} set and VOC12 \emph{trainval} set as the training set with 22,136 images in total, which is used for the omni-supervision experiments. The model is evaluated on the VOC07 \emph{test} set. 

\vspace{-4mm}
\paragraph{COCO~\cite{lin2014microsoft}} COCO \emph{train2017} set of 118,287 images is used as the training set for splitting. The model is evaluated on the COCO \emph{val2017} set.

\vspace{-4mm}
\paragraph{Objects365~\cite{shao2019objects365}} To have faster experiments, 93,455 images
are sampled from the Objects365 official training set as the training set for the omni-supervision experiments. In the process, since this dataset is long-tailed, we ensure that there is at least one image per category. Performance is evaluated on the official validation set.

The cost number (the second column from right) in Table \ref{tab:omni-per} is computed by considering the mixture ratio, the dataset size and the cost per image in Table \ref{tab:cost}. For example, for the first row of Bees, the cost is
$25 = (3596*0.05*249.9+3596*0.8*6.1+3596*0.15*50)/3600$.

\begin{figure*}
\begin{minipage}{1.0\textwidth}
      \begin{subfigure}[b]{0.5\textwidth}
      \captionsetup{width=.8\linewidth}
             \includegraphics[width=\linewidth]{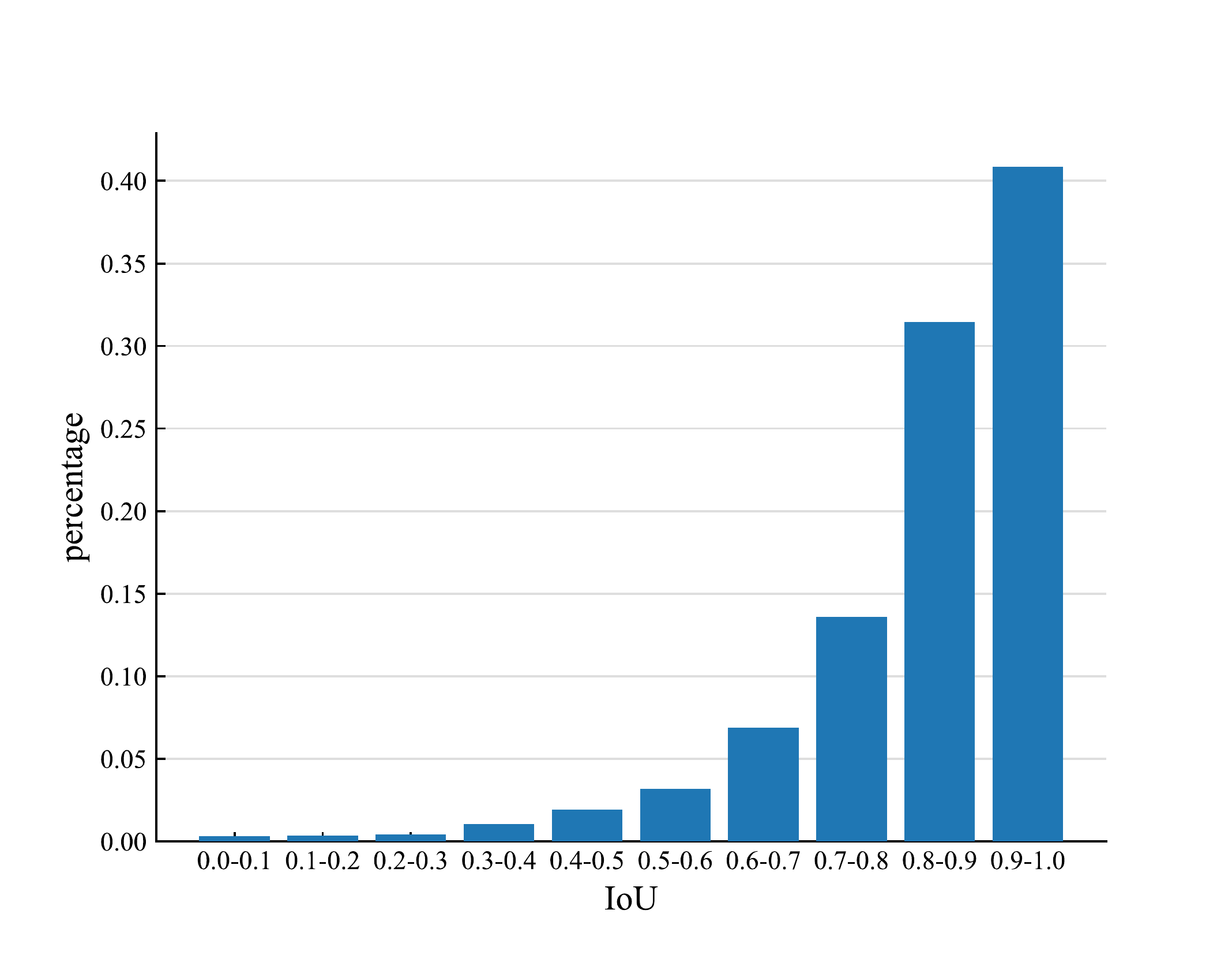}
              \caption{Extreme Clicking on VOC. mean is 0.83; std is 0.15; second-order moment is 0.02.}
                \label{fig:voc}
        \end{subfigure}%
        \begin{subfigure}[b]{0.5\textwidth}
        \captionsetup{width=.8\linewidth}
             \includegraphics[width=\linewidth]{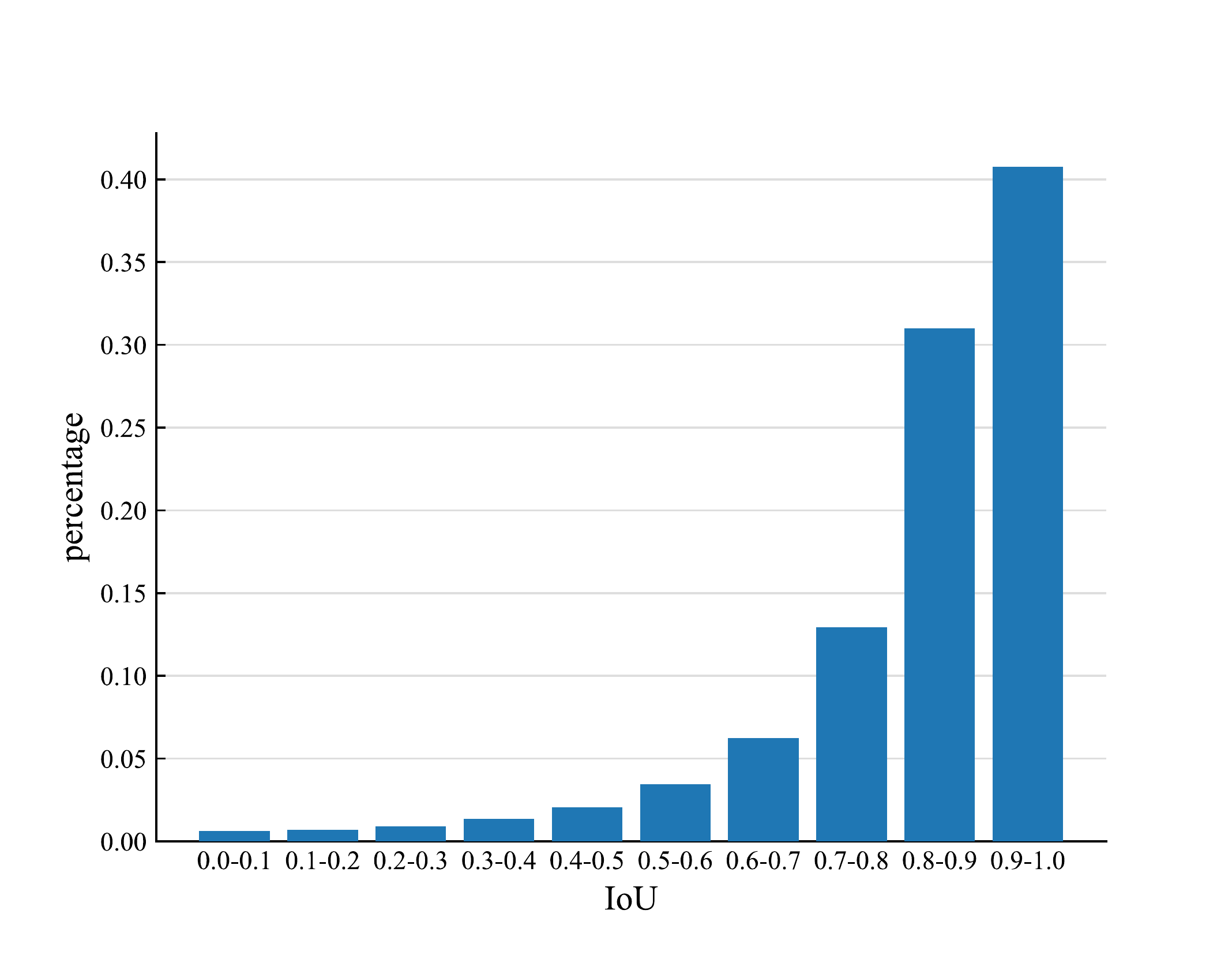}
                \caption{Simulated Extreme Clicking on COCO. mean is 0.82; std is 0.16; second-order moment is 0.02.}
                \label{fig:coco}
        \end{subfigure}%
        \caption{The distribution of mIoU between the BoxesEC and ground truth.}\label{fig:simulation}
\end{minipage}
\end{figure*}

\subsection{The Simulation of Extreme Clicking Boxes}

Because Extreme Clicking~\cite{papadopoulos2017extreme} does not release the annotations except for VOC~\cite{everingham2010pascal}, we simulate the boxes generated by Extreme Clicking for the other four datasets in our experiments. In detail, for each dataset, Gaussian noise is added to the ground truth bounding box coordinates, such that the distribution of mean Intersection over Union (mIoU) between the simulated boxes and ground truth boxes is close to the mIoU distribution between the given Extreme Clicking boxes and the ground truth boxes on VOC. The value of mIoU can be controlled by varying the covariance matrix of the Gaussian noise. 
Figure \ref{fig:simulation} shows the comparison of mIoU distribution of Extreme Clicking (left) and our simulation on COCO $10\%$Fully+$90\%$BoxesEC setting  (right) as an example. Their statistics comparison is: 1) Mean: 0.83 v.s. 0.82; Std: 0.15 v.s. 0.16; Second-order moment: 0.02 v.s. 0.02. These have shown our simulation is close to Extreme Clicking.

\subsection{Other Implementation and Training Details}

The number of epoch for Burn-In stage depends on the size of the labeled data in our experiments. The total epoch number is chosen until the training saturated. They are shown in Table \ref{tab:train-detail}. All models of Deformable DETR are trained with total batch size of 16. For other hyperparameters, we mainly follow the settings of Unbiased Teacher~\cite{liu2021unbiased} and Deformable DETR~\cite{zhu2020deformable}. For example, in (\ref{equ:loss1}), weight $\alpha=2$, $\beta=5$ to balance the classification loss ($\mathcal{L}^{cls}$ ) and regression loss ($\mathcal{L}^{\text{box}}$). $\mathcal{L}^{cls}$ is the focal loss with default hyperparameters,
and $\mathcal{L}^{\text{box}}=2\mathcal{L}_{\text{iou}}+5\mathcal{L}_{\text{L1}}$ combines generalized IoU loss and L1 loss. EMA smoothing constant $k=0.9996$ in (\ref{equ:L_t}). Weights $\lambda_{\text{iou}}=2$, $\lambda_{L1}=5$ in (\ref{equ:box_gamma}) for box matching. The object query number is $K=300$ in Section \ref{sec:filtering}.

\begin{table}
\setlength{\abovecaptionskip}{-2.0pt}
\setlength{\tabcolsep}{2pt}
\small
\begin{center}
\begin{tabular}{l|c|cc}
Dataset & Fully ($\%$) & Burn-In Epochs &Total Epochs \\
\thickhline
\multirow{4}{*}{Bees} &  5& 600& 1000\\
&   10& 400 & 1000\\ 
&   20& 200 & 1000\\ 
&  30 & 100 & 1000 \\ \cline{2-4}
\hline
\multirow{4}{*}{CrowdHuman} &  5& 400& 800\\
&10& 200 & 500\\ 
&20& 100 & 500\\ 
&   30 & 80 & 500\\ \cline{2-4}
\hline
\multirow{4}{*}{VOC} & 8& 300&800 \\
&   10 & 200 & 500\\ 
&   20 & 100 & 500\\ 
&     30& 80& 200\\ \cline{2-4}
\hline
\multirow{6}{*}{COCO} &  1 & 400& 800\\
&     5 &80& 500 \\
&     8 &60& 300 \\
&     10 &40& 200 \\
&     20 &20& 150 \\ 
&    30  & 15 &100\\ \cline{2-4}
\hline
\multirow{5}{*}{Objects365} & 8&130 & 300 \\
&    10&100 & 200 \\ 
&    20  & 50&200 \\
&    25  & 40&200 \\
&    30  & 30&200 \\
\hline
\end{tabular}
\end{center}
\caption{The epoch setting details of omni-supervision training}
\label{tab:train-detail}
\end{table}

\end{document}